\RequirePackage{fix-cm}
\documentclass[twocolumn,natbib]{svjour3}          
\smartqed  
\usepackage{graphicx}
\usepackage[colorlinks,bookmarksopen,bookmarksnumbered,linkcolor=blue,citecolor=cyan,urlcolor=cyan]{hyperref}

\usepackage{adjustbox}
\usepackage{multirow}
\usepackage[table,xcdraw]{xcolor}
\usepackage{moreverb,url}
\usepackage{booktabs}
\usepackage{breakcites}

\renewcommand{\cite}{\citep}

\renewcommand{\endnote}{\footnote}

\graphicspath{{./pictures/}}

\DeclareGraphicsExtensions{.pdf,.jpeg,.png,.eps}

\journalname{ }
\begin{document}

\title{Are we done with object recognition? The iCub robot's perspective.
}



\author{Giulia Pasquale \and Carlo Ciliberto \and Francesca Odone \and Lorenzo Rosasco \and Lorenzo Natale 
}


\institute{G. Pasquale \and L. Natale
\at Humanoid Sensing and Perception, Istituto Italiano di Tecnologia, Via Morego 30, Genova, IT\\
 \email{giulia.pasquale@iit.it} 
\and
G. Pasquale \and F. Odone \and L. Rosasco
\at Dipartimento di Informatica, Bioingegneria, Robotica e Ingegneria dei Sistemi, Universit\'a degli Studi di Genova, Via Dodecaneso 35, Genova, IT
\and
G. Pasquale \and L. Rosasco
\at Laboratory for Computational and Statistical Learning (IIT@MIT), Massachusetts Institute of Technology, Bldg. 46-5155, 43 Vassar Street, Cambridge, MA
\and
L. Rosasco
\at Center for Brains, Minds and Machines, Massachusetts Institute of Technology, 77 Massachusetts Avenue, Cambridge, MA
}


\maketitle

\begin{abstract}
We report on an extensive study of the benefits and limitations of current deep learning approaches to object recognition in robot vision scenarios, introducing a novel dataset used for our investigation. To avoid the biases in currently available datasets, we consider a natural human-robot interaction setting to design a data-acquisition protocol for visual object recognition on the iCub humanoid robot. Analyzing the performance of off-the-shelf models trained off-line on large-scale image retrieval datasets, we show the necessity for knowledge transfer. We evaluate different ways in which this last step can be done, and identify the major bottlenecks affecting robotic scenarios. By studying both object categorization and identification problems, we highlight key differences between object recognition in robotics applications and in image retrieval tasks, for which the considered deep learning approaches have been originally designed. In a nutshell, our results confirm the remarkable improvements yield by deep learning in this setting, while pointing to specific open challenges that need be addressed for seamless deployment in robotics. 

\keywords{Humanoid Robotics \and Robot Vision \and Visual Object Recognition \and Machine Learning \and Deep Learning \and Transfer Learning \and Image Dataset \and Dataset Collection \and Representation Invariance \and iCub}
\end{abstract}


\section{Introduction}

Artificial intelligence has recently progressed dramatically, largely thanks to the advance in deep learning. Computational vision, specifically object recognition, is perhaps the most obvious example where deep learning has achieved so stunning results to raise the question of whether this problem is actually solved~\cite{krizhevsky2012,simonyan2014,szegedy2015,he2015,he2016}. 
Should this be the case, robotics would be a main field where the benefits could have far reaching effect. Indeed, the lack of reliable visual skills is largely considered a major bottleneck for the successful deployment of robotic agents in everyday life~\cite{kemp2007}.

With this perspective in mind, we have recently started an effort to isolate and quantify the benefits and limitations, if any, of deep learning approaches to visual object recognition in robotic applications~\cite{pasquale2015,pasquale2016iros}. The remarkable performance of deep learning methods for object recognition has in fact been primarily reported on computer vision benchmarks such as~\cite{griffin2007,everingham2010,everingham2015,russakovsky2015}, which are essentially designed for large-scale image retrieval tasks and are hardly representative of a robotic application setting (a motivation common to other recent works such as~\cite{pinto2011, leitner2015, oberlin2015, borji2016}).

Clearly, visual perception is only one of the possible sensory modalities equipping modern robots, that can be involved in the object recognition process (see for example~\cite{luo2017,higy2016}). In addition it has been shown that the physical interaction with the environment can be used to aid perception~\cite{pinto2016}, demonstrating that there is more than ``just'' vision to object recognition.
Nonetheless, visual recognition of objects is evidently the very first and critical step for autonomous agents to act in the real world. Current deep learning-based artificial systems perform so well in this task, that it seems natural to ask how far they can go, before further perceptual cues are needed. To this end, in this work we focus on the problem of object recognition in robotics using only visual cues.

We consider a prototypical robotic scenario, where a humanoid robot is taught to recognize different objects by a human through natural interaction. We started with the design of a dataset tailored to reflect the visual ``experience'' of the robot in this scenario. This dataset is rich and easy to expand to include more data and complex perceptual scenarios. It includes several object categories with many instances per category, hence allowing to test both object categorization and identification tasks. Notably, the dataset is segmented into multiple sets of image sequences per object, representing specific viewpoint transformations like scaling, in- and out-of-plane rotations, and so forth. It provides a unique combination that, to our knowledge, was missing in the literature, allowing for articulated analyses of the robustness and invariance properties of recognition systems within a realistic robotic scenario. Since we used the iCub robot~\cite{metta2010}, we called it iCubWorld Transformations (iCWT for short).

We performed extensive empirical investigation using different state-of-the-art Convolutional Neural Network architectures, demonstrating that off-the-shelf models do not perform accurately enough and pre-trained networks need to be adapted (fine-tuned) to the data at hand to obtain substantial improvements. However, these methods did not quite provide the close to perfect accuracy one would wish for. We hence proceeded taking a closer look at the results, starting from the question of whether the missing gap could be imputed to lack of data. Investigating this latter question highlighted a remarkable distance between iCWT and other datasets such as ImageNet~\cite{russakovsky2015}. We identified clear differences between the object recognition task in robotics with respect to scenarios typically considered in learning and vision. In fact, as we show in the paper, our empirical observations on iCWT are extended to other robotic datasets like, e.g., Washington RGB-D~\cite{lai2011}.

Along the way, our analysis allowed also to test the invariance properties of the considered deep learning networks and quantify their merits not only for categorization but also for identification. 

The description and discussion of our empirical findings is concluded with a critical review of some of the main venues of improvements, from a pure machine learning perspective but also taking extensive advantage of the robotic platform. Indeed, bridging the gap in performance appears to be an exciting avenue for future multidisciplinary research. 

In the next Section we discuss several related works, while the rest of the paper is organized as follows: Sec.~\ref{sec:icubworld} introduces the iCWT dataset and its acquisition setting. In Sec.~\ref{sec:methods} we review the deep learning methods considered for our empirical analysis, which is reported in Sec.~\ref{sec:experiments} for the categorization task and in Sec.~\ref{sec:identification} for object identification. In Sec.~\ref{sec:rgbd} we finally show how the same observations drawn from experiments on iCWT hold also for other robotic datasets, specifically we consider Washington RGB-D~\cite{lai2011}. Sec.~\ref{sec:discussion} concludes our study with the review of possible directions of improvement for visual recognition in robotics.

\subsection{Deep Learning for Robotics} 

\begin{figure*}[t]
\centering
\includegraphics[width=1.85\columnwidth]{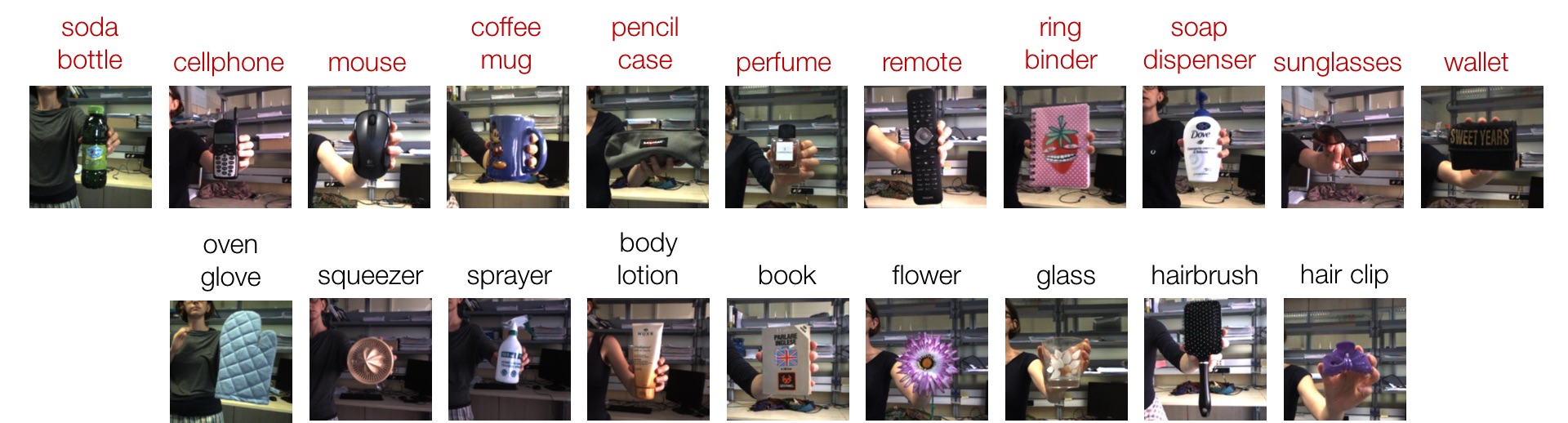}
\caption{{\bf iCWT categories}. For each category in iCWT, we report one example image for one instance. (Red) categories appear also in the ILSVRC 2012 dataset. (Black) categories appear in ImageNet but not in ILSVRC (see supplementary material for more details).}
\label{fig:20objects}
\end{figure*}

Deep Learning methods are receiving growing attention in robotics, and are being adopted for a variety of problems such as object recognition~\cite{schwarz2015, eitel2015, pinto2016, held2016, pasquale2016iros}, place recognition and mapping~\cite{sunderhauf2015, sunderhauf2016}, object affordances~\cite{nguyen2016}, grasping \cite{redmon2015,pinto2015,levine2016a} and tactile perception~\cite{baishya2016}. We limit our discussion to the work on object recognition, which is more relevant to this paper.

In~\cite{schwarz2015, eitel2015} the authors demonstrate transfer learning from pre-trained deep Convolutional Neural Networks (CNNs) and propose a way to include depth information from an RGB-D camera by encoding depth data into RGB with colorization schemes. The main idea of~\cite{schwarz2015} is to extract a feature vector from the CNN and train a cascade of Support Vector Machines (SVMs) to discriminate the object's class, identity and position, while~\cite{eitel2015} proposes a pipeline to fine-tune the CNNs weights on RGB-D input pairs. Both works focus on the Washington RGB-D benchmark~\cite{lai2011} and improve its state-of-the-art performance.
In this paper, we consider a less constrained setting, in that CNNs are trained with data acquired by the robot during natural interaction (which undergo, therefore, more challenging viewpoint transformations). We adopt similar techniques for transfer learning, but we assess a wide range of architectures and fine-tuning approaches. 

The work in~\cite{pinto2016} shows how the robot can use self-generated explorative actions (like pushing and poking objects) to autonomously extract example data and train a CNN. In contrast to~\cite{pinto2016}, the use of a human teacher (see Sec.~\ref{sec:icubworld} for details on the acquisition setup) gives us more control on the object transformations. On the other hand, our work could be extended by introducing self-supervision using explorative actions similar to the ones in~\cite{pinto2016}. 

The work in~\cite{held2016} and our work in~\cite{pasquale2016iros} are closely related to the work presented in this paper in that they investigate invariance properties of CNNs and how to improve them in order to learn from few examples. They focus, however, on instance recognition, whereas in this paper we include in the analysis -- thus significantly extending~\cite{pasquale2016iros} -- the problem of object categorization. In addition, we perform a detailed investigation of various transfer learning approaches and present a systematic evaluation of the recognition performance for specific viewpoint transformations. 

\subsection{Datasets for Visual Recognition in Robotics}

In the literature, several datasets have been used to benchmark visual object recognition in robotics: COIL~\cite{nene96}, ALOI~\cite{geusebroek2005}, Washington RGB-D~\cite{lai2011}, KIT~\cite{kasper2012}, SHORT-100~\cite{rubio2014}, BigBIRD~\cite{singh2014}, Rutgers Amazon Picking Challenge RGB-D Dataset~\cite{rennie2016} are only some examples. One of the main characteristic of these datasets is to capture images of an object while it undergoes viewpoint transformations. However, these datasets are usually acquired in strictly controlled settings (i.e. using a turntable),
because they are aimed to provide also pose annotations. As a consequence, image sequences differ from the actual data that can be gathered by a robot in its operation: they do not show substantial variations of objects' appearance and often background or light changes are missing or under-represented. See~\cite{leitner2015} for a review of the major limitations of current datasets.

The NORB dataset~\cite{lecun2004}, albeit acquired in a similar turntable setting, is one of the first benchmarks released in support of the investigation of invariance properties of recognition methods. 
A similar study focusing on deep learning methods is described in~\cite{borji2016} using the iLab-$20$M dataset. This aims to be a comprehensive benchmark for visual object recognition that, while representing a high number of object instances, provides also varied images of each object. While presenting a remarkably higher variability, however, also iLab-$20$M is acquired in a turntable setting (to collect pose annotations), hence suffering from similar limitations. 

Our iCWT dataset separates from previous work in that objects are captured during ``natural'' transformations. Acquisition is performed in a ``semi-controlled'' setting intended to reproduce typical uncertainties faced by the visual recognition system of a robot during a real-world task. Very few works in the literature consider ``real'' (i.e. non-synthetic)  transformations when evaluating the invariance properties of visual representations (see, e.g.,~\cite{goodfellow2009}). To our knowledge, iCWT is the first dataset to address invariance-related questions in robotics. Moreover, it accounts for a much wider range of visual transformations with respect to previous datasets. 
In the following, we discuss the data collection and the acquisition setting in detail. Note that, while we used an initial subset of iCWT in~\cite{pasquale2016iros}, in this paper we present the dataset for the first time in its entirety.

\begin{figure}[t]
\centering
\includegraphics[width=\columnwidth]{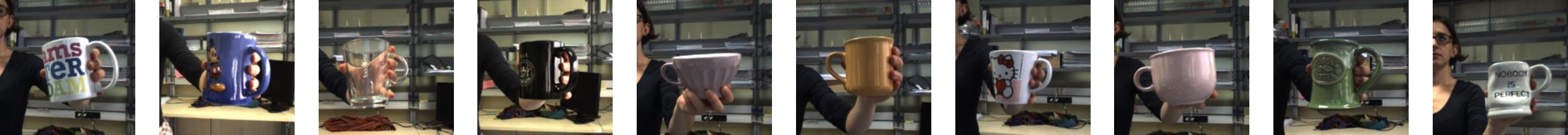}
\caption{{\bf Semantic variability.} Sample images for the different object instances in the {\em mug} category to provide a qualitative intuition of the semantic variability in iCubWorld. See Fig.~\ref{fig:200objects} in the supplementary material for more examples.}
\label{fig:10mugs}
\end{figure}


\section{The iCubWorld Transformations Dataset}
\label{sec:icubworld}

The iCubWorld Transformations (iCWT) is a novel benchmark for visual object recognition, which we use for the empirical analysis in this work. iCWT is the latest release within the iCubWorld\endnote{\url{https://robotology.github.io/iCubWorld/}} project, whose goal is to benchmark and improve artificial visual systems for robotics. iCubWorld datasets~\cite{fanello2013icubworld,pasquale2015} are designed to record a prototypical visual ``experience'' of a robot while it is performing vision-based tasks. To this end, we devised and implemented a simple human-robot interaction application, during which we acquire images for the dataset from the robot's cameras.

There is a remarkable advantage in collecting iCubWorld directly from the robot platform. The resulting data collection offers a natural testbed, as close as possible to the real application. This ensures that the performance measured {\em off-line} on iCubWorld can be expected to generalize well when the system is deployed on the actual robot. Note that this aspect of iCubWorld is extremely relevant since visual biases make it typically difficult to generalize performances across different datasets and applications, as already well known from previous work~\cite{pinto2008,torralba2011,khosla2012,hoffman2013,rodner2013,model2015, stamos2015,tommasi2015} and also shown empirically in Sec.~\ref{sec:ots} of this paper.

Currently, to acquire iCubWorld releases we did not make extensive use of the robot's physical capabilities (e.g., manipulation, exploration, etc.). This was done because latest deep learning methods already achieve remarkable performance by relying solely on visual cues and our goal was to evaluate their accuracy in {\em isolation} in a robotic setting. While exploiting the robot body could provide further advantages to modeling and recognition (see for instance~\cite{moldovan2012},~\cite{leitner2014} and~\cite{dansereau2016}), it would also prevent us to assess the contribution of the visual component alone.

\subsection{Acquisition Setup}
\label{sec:acquisitionsetup}

During data acquisition a human ``teacher'' shows an object to the robot and pronounces the associated label. The robot exploits bottom-up visual cues to track and record images of the object while the human moves and shows it from different poses. Annotations are automatically recorded for each image in terms of the object's label (provided by the human) and bounding box (provided by the tracker).

Differently from all previous releases of iCubWorld, for the acquisition of iCWT we decided to rely on a tracker based on depth segmentation~\cite{pasquale2015depth} instead of one based on the detection of object motion, because this greatly simplifies the acquisition while providing more stable and precise bounding boxes.

\begin{figure}[t]
\centering
\includegraphics[width=\columnwidth]{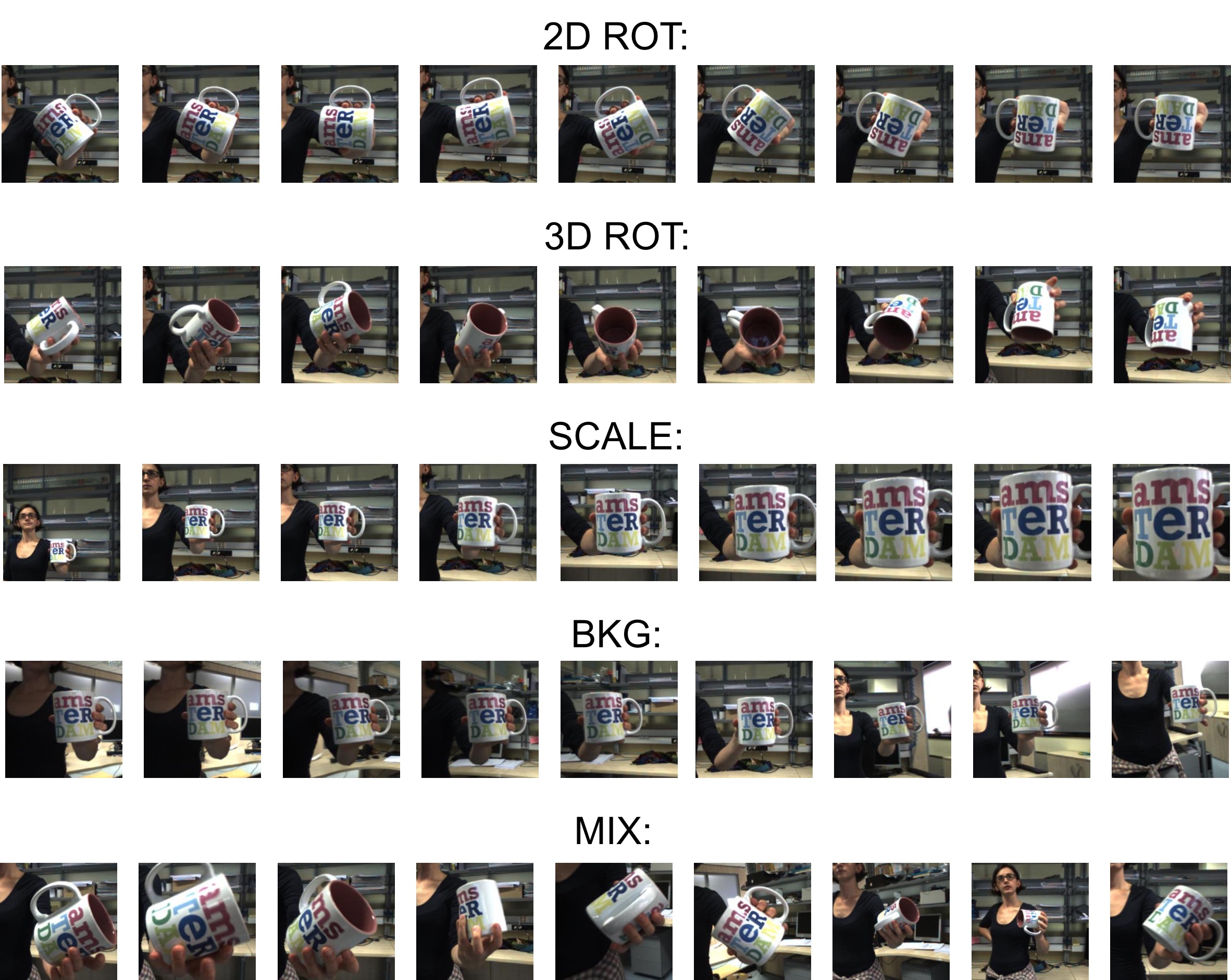}
\caption{{\bf Visual transformations.} Excerpts from the sequences acquired for one mug, representing the object while it undergoes specific visual transformations.}
\label{fig:5transf}
\end{figure}

We developed an application to scale the acquisition seamlessly to hundreds of objects, which were collected during multiple interactive sessions. iCWT is available on-line and we plan to make also this application publicly available in order for other laboratories to use the same protocol to collect their own (or possibly contribute to) iCubWorld. 
The proposed acquisition approach allows to build large-scale data collections, fully annotated through natural interaction with the robot and with minimal manual intervention. Moreover, the application can be directly deployed on other iCub robots and, being relatively simple, can also be adapted to other platforms.

\begin{table}[t]
\centering
\caption{Summary of the {\it iCubWorld Transformations} dataset}
\begin{adjustbox}{max width=\columnwidth}
\begin{tabular}{ccccc}
\multirow{2}{*}{\bf\# Categories}   & \bf \# Obj. per & \multirow{2}{*}{\bf \# Days} & \multirow{2}{*}{\bf Transformations} & \bf \# Frames per \\
                                    & \bf Category    & & & \bf Session \\
\toprule 

\multirow{3}{*}{$20$} & \multirow{3}{*}{$10$} & \multirow{3}{*}{$2$} & $2D$ ROT, $3D$ ROT & \multirow{2}{*}{$150$} \\
                      &                       &                      & SCALE, BKG         & \\[0.4ex]
                      &                       &                      & MIX                & $300$

\end{tabular}
\end{adjustbox}
\label{tab:icubworld}
\end{table}

\subsection{Dataset Overview}
\label{sec:overview}

iCWT includes $200$ objects evenly organized into $20$ categories that can be typically found in a domestic environment. Fig.~\ref{fig:20objects} reports a sample image for each category in iCWT: $11$ categories (in red in the figure) are also in the ImageNet Large-Scale Visual Recognition Challenge (ILSVRC) 2012~\citep{russakovsky2015}, i.e. we found semantically and visually similar classes among the $1000$ of the classification challenge. The remaining $9$ categories do not appear ILSVRC but belong (or are similar) to a synset in the larger ImageNet dataset~\cite{deng2009}. To provide a qualitative intuition of the semantic variability within a given category, namely the different visual appearance of object instances in the same category, Fig.~\ref{fig:10mugs} shows a sample image from the $10$ instances in the {\it mug} category. We refer the reader to the supplementary material (Fig.~\ref{fig:200objects}) for example images of all object instances in iCWT.

In iCWT each object is shown in multiple image sequences while it undergoes specific visual transformations (such as rotations, scaling or background changes).
For each object instance, we acquired $5$ image sequences, while the human supervisor was performing a different visual transformation on the object. Fig.~\ref{fig:5transf} reports excerpts of these sequences, which contain, respectively:

\begin{description}[noitemsep]

\item[\bf $2D$ Rotation] The human rotated the object parallel to the camera image plane. The same scale and position of the object were maintained (see Fig.~\ref{fig:5transf}, first row).
\item[\bf $3D$ Rotation] Similarly to $2D$ rotations, the object was kept at same position and scale. However, this time the human applied a generic rotation to the object (not parallel to the image plane). As a consequence different ``faces'' of the object where shown to the camera (Fig.~\ref{fig:5transf}, second row).
\item[\bf Scale] The human moved the object towards the cameras and back, thus changing the object's scale in the image. No change in the object orientation (no $2D$ or $3D$ rotation) was applied (Fig.~\ref{fig:5transf}, third row).
\item[\bf Background] The human moved the object around the robot, keeping approximately the same distance (scale) and pose of the object with respect to the camera plane. Because the robot tracks the object, the background changes while the object appearance remains approximately the same (Fig.~\ref{fig:5transf}, fourth row).
\item[\bf Mix] The human moved the object freely in front of the robot, as a person would naturally do when showing a new item to a child. In this sequence all nuisances in all combinations can appear (Fig.~\ref{fig:5transf}, fifth row).

\end{description}

Each sequence is composed by approximately $150$ images acquired at $8$ frames per second in the time interval of $20s$, except for the {\it Mix} sequence that lasted $40s$ and comprises $\sim300$ images. 
As anticipated, the acquisition of the $200$ objects was split into multiple sessions performed in different days. The acquisition location was always the same (with little uncontrolled changes in the setting across days). 
The illumination condition was not artificially controlled, since we wanted to investigate its role as a further nuisance: to this end, we acquired objects at different times of the day and in different days, so that lighting conditions are slightly changing across the $200$ objects (but not within the five sequences of an object, which were all acquired in the span of few minutes). Moreover, we repeated the acquisition of each object in two different days, so that we ended up with $10$ sequences per object, containing $5$ visual transformations in $2$ different illumination conditions.
The adopted iCub's cameras resolution is $640 \times 480$.
Both left and right images provided by the iCub's stereo pair were acquired, to allow for offline computation of the disparity map and possibly further improvement of the object's localization and segmentation.
We recorded the centroid and bounding box of the object provided by the tracker at each frame. Tab.~\ref{tab:icubworld} summarizes the main characteristics of iCubWorld Transformations. We refer to the iCub's website~\footnote{\url{http://www.icub.org/}} for details about the cameras and their setting.


\begin{figure*}
\centering
\includegraphics[width=0.8\textwidth]{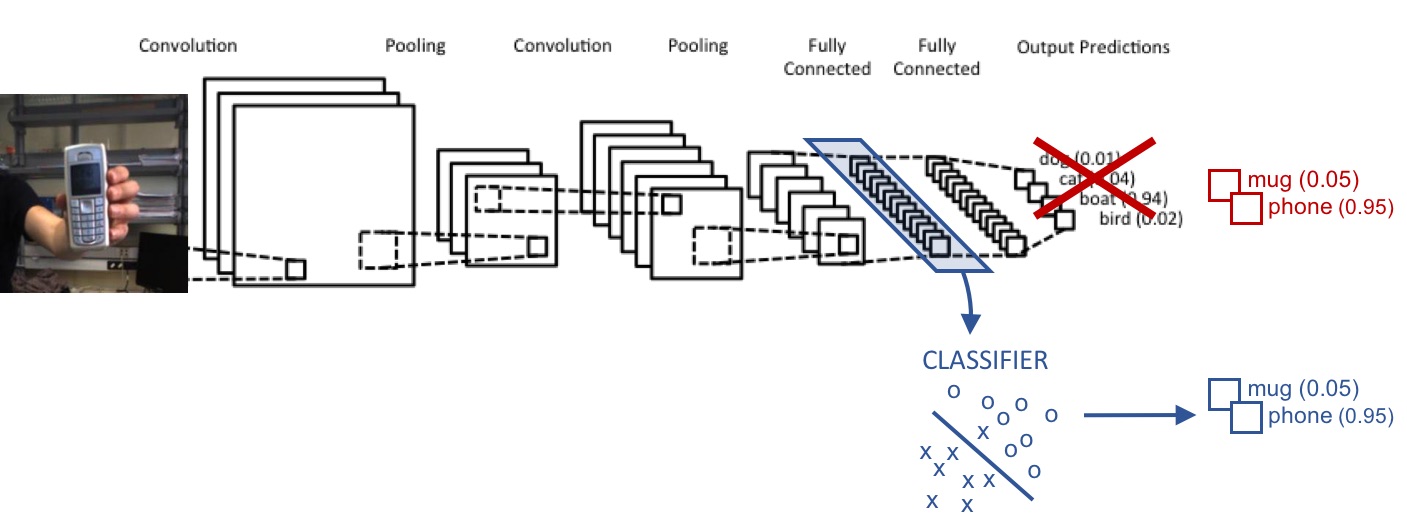}
\begin{minipage}{\linewidth}
\caption[a]{Example of a Convolutional Neural Network (Sec.~\ref{sec:deepCNNs}) and of the two knowledge transfer approaches considered in this work (Sec.~\ref{sec:transferlearning}). {\bf (Blue Pipeline) Feature Extraction}: in this case the response of one of the layers is used as a feature vector for a ``shallow'' predictor like RLSCs, SVMs (see Sec.~\ref{sec:feature-extraction}), which is trained on the new task. {\bf (Red Pipeline) Fine-tuning}: in this case the network is trained end-to-end to the new task by replacing the final layer and using the original model as a ``warm-restart'' (see Sec.~\ref{sec:fine-tuning}). Network image from~\footnotemark.}
\label{fig:cnn}
\end{minipage}
\end{figure*}

\section{Methods}
\label{sec:methods}

\subsection{Deep Convolutional Neural Networks}
\label{sec:deepCNNs}

Deep Convolutional Neural Networks (CNNs) are hierarchical models iterating multiple processing layers. Their structure aims to map the input (say, an image) into a series of {\em feature maps} or {\em representations} that progressively select the visual features which are most relevant to the considered task. The prototypical structure of a CNN (see Fig.~\ref{fig:cnn}\footnotetext{\url{https://www.clarifai.com/technology}}) alternates blocks of (i) convolution (followed by element-wise non linearities, as sigmoids~\cite{bishop2006} or ReLUs~\cite{he2015}), and (ii) spatial pooling and downsampling. The convolution (plus non linear) layers are such to progressively extract, from the image, maps of selected and more complex features (from edges to local pattern up to object parts), which are relevant to the task at hand~\cite{chatfield2014,yosinski2014,donahue2014,zeiler2014}. Spatial downsampling and pooling layers make these features more robust (ideally, invariant) to transformations of the input at increasingly larger scales. A common strategy in image classification is to follow these blocks with one or more fully connected layers (namely, a standard Neural Network). In classification settings, the last layer is a {\em softmax} function, which maps the output into class-likelihood scores, whose maximum is the predicted class.

The modular structure of CNNs allows training their parameters (namely, convolution filters) simultaneously for all layers (also known as {\em end-to-end learning}) via back-propagation~\cite{lecun89}. Given the large number of parameters to be optimized (in the order of millions), CNNs typically need large amounts of training data to achieve good performance. Often, the training examples are artificially increased by synthetically modifying the images ({\em data augmentation}). To further mitigate the risk of overfitting, regularization techniques such as L$2$ regularization, dropout~\citep{hinton2012,srivastava2014} or, more recently, batch normalization~\citep{ioffe2015}, have proved helpful.

In this work we investigate the performance of modern CNNs on the robotic setting of iCubWorld. To this end, we selected four architectures achieving the highest accuracy on the ImageNet Large-Scale Visual Recognition Challenge (ILSVRC)~\cite{russakovsky2015}. We used their implementation trained on ILSVRC $2012$ and publicly available within the {\sc Caffe}~\citep{jia2014} framework. Specifically we consider {\bf CaffeNet}\footnotemark\footnotetext{\url{https://github.com/BVLC/caffe/tree/master/models/bvlc\_reference\_caffenet}}, a variation of AlexNet~\citep{krizhevsky2012}, and {\bf VGG-16}\footnotemark\footnotetext{\url{http://www.robots.ox.ac.uk/$\sim$vgg/research/very\_deep/}}~\citep{simonyan2014}. These two models concatenate a number of convolution layers ($5$ and $13$) with $3$ fully connected layers and comprise around $60$ and $140M$ parameters respectively.
Then we consider {\bf GoogLeNet}\footnotemark\footnotetext{\url{https://github.com/BVLC/caffe/tree/master/models/bvlc\_googlenet}}~\citep{szegedy2015} and {\bf ResNet-50}\footnotemark\footnotetext{\url{https://github.com/KaimingHe/deep-residual-networks}}~\citep{he2016}, which slightly diverge from the standard CNN structure described above in that they respectively employ so-called {\it inception modules} 
or {\it residual connections}, and the number of parameters is also reduced ($4M$ for GoogLeNet, $22$ layers, and $20M$ for ResNet-50, $50$ layers).

\subsection{Transfer Learning Techniques}
\label{sec:transferlearning}

The need for large datasets to successfully train deep CNNs could in principle prevent their applicability to problems where training data is scarce. However, recent empirical evidence has shown that the knowledge learned by a CNN on a large-scale problem can be ``transferred'' to multiple domains. 
In this section we review two of the most well-established methods which we empirically assess in our experiments, namely {\em feature extraction} and {\em fine-tuning}.

\begin{table}[t]
\centering
\caption{Feature extraction layers for the four architectures considered in this work. We used the notation adopted in {\sc Caffe}, in which the number identifies the layer number and the label specifies its type (i.e., fully connected or pooling layer).}
\label{tab:feat_extraction}
\footnotesize
\begin{tabular}{r l}
\textbf{Model} & \textbf{Output Layer} \\
\toprule
CaffeNet       & fc6 or  fc7                   \\
GoogLeNet      & pool5/7x7\_s1             \\
VGG-16         &  fc6 or  fc7                  \\
ResNet-50      & pool5 \\
\bottomrule                       
\end{tabular}
\end{table}

\subsubsection{Feature Extraction}
\label{sec:feature-extraction}

It has been shown that CNNs trained on large, varied image datasets can be used on smaller datasets as generic ``feature extractors'', leading to remarkable performance~\cite{schwarz2015,oquab2014,sunderhauf2015,pasquale2016iros}. This is typically done by training a classifier (such as a Support Vector Machine (SVM) or a Regularized Least Squares Classifier (RLSC)~\cite{bishop2006}) on feature vectors obtained as the activations of intermediate network layers. This strategy, depicted in Fig.~\ref{fig:cnn} (Blue pipeline), can also be interpreted as changing the last layer of the CNN and training it on the new dataset, while keeping fixed all other network parameters.\\

\noindent{\bf Implementation Details}. We used this strategy to transfer knowledge from CNNs trained on ImageNet to iCWT. The CNN layers used for our experiments are reported in Tab.~\ref{tab:feat_extraction} for the four architectures considered in the paper. We used RLSC with a Gaussian Kernel as classifier for the CNN-extracted features. In particular we implemented the Nystr\"om sub-sampling approach proposed in~\citep{rudi2015, rudi2016}, which is computationally appealing for mid and large-scale settings and significantly speeded up our experimental evaluation. We refer to the supplementary material for details on model selection and image preprocessing operations.

\subsubsection{Fine-tuning}
\label{sec:fine-tuning}

\begin{table}[t]
\centering
\caption{Fine-tuning protocols for {\it CaffeNet} and {\it GoogLeNet}. Base LR is the starting learning rate of all layers that are initialized with the original model. The FC layers that are learned from scratch are indicated using their names in {\sc Caffe} models ($2^{nd}$ row), specifying the starting learning rate used for each of them. For the other parameters, we refer the reader to {\sc Caffe} documentation.}
\label{tab:tuningStrategies}
\begin{adjustbox}{max width=\columnwidth}
\begin{tabular}{c c c c c}
& \multicolumn{2}{c}{\bf CaffeNet} & \multicolumn{2}{c}{\bf GoogLeNet} \\ 
 & {\it adaptive}  & {\it conservative} & {\it adaptive} & {\it conservative} \\
\toprule
{\bf Base LR}          & 1e-3  & 0  & 1e-5 & 0 \\
\midrule
 & & fc8: 1e-4 & \multicolumn{2}{c}{loss3/classifier: 1e-2} \\
 \multirow{-2}{*}{ {\bf Learned}}& & fc7: 1e-4 & \multicolumn{2}{c}{loss1(2)/classifier: 1e-3} \\
 \multirow{-2}{*}{ {\bf FC Layers}} & \multirow{-3}{*}{fc8: 1e-2} & fc6: 1e-4 &  \multicolumn{2}{c}{loss1(2)/fc: 1e-3} \\
\midrule
&   \multicolumn{2}{c}{fc7: 50} &  \multicolumn{2}{c}{pool5/drop\_7x7\_s1: 60} \\
\multirow{-2}{*}{ {\bf Dropout (\%)} } &  \multicolumn{2}{c}{fc6: 50}  &  \multicolumn{2}{c}{loss1(2)/drop\_fc: 80} \\
\midrule
{\bf Solver} & \multicolumn{2}{c}{SGD} & \multicolumn{2}{c}{Adam} \\
{\bf LR Decay Policy } & \multicolumn{2}{c}{Polynomial (exp 0.5)} & \multicolumn{2}{c}{No decay} \\
\midrule
\# {\bf Epochs}  & 6 & 36 & \multicolumn{2}{c}{6} \\
\midrule
{\bf Batch Size } &  \multicolumn{2}{c}{256}  &  \multicolumn{2}{c}{32} \\
\bottomrule                                      
\end{tabular}
\end{adjustbox}
\end{table}

A CNN trained on a large image dataset can also be used to ``warm-start'' the learning process on other (potentially smaller) datasets. This strategy, known as {\em fine-tuning}~\citep{chatfield2014,simonyan2014}, consists in performing back-propagation on the new training set by initializing the parameters of the network to those previously learned (see Fig.~\ref{fig:cnn} (Red pipeline)). In this setting it is necessary to adapt the final layer to the new task (e.g., by changing the number of units in order to account for the new number of classes to discriminate). 

A potential advantage of fine-tuning is that it allows to adapt the parameters of all layers to the new problem (rather than only those in the final layer). This flexibility however comes at the price of a more involved training process: the choice of the (many) hyper-parameters available (e.g., which layers to adapt and how strongly) can have a critical impact on performance, especially when dealing with large architectures and smaller training sets. Fine-tuning has been recently used to transfer CNNs learned on ImageNet to robotic tasks (see, e.g., \cite{eitel2015,pinto2015, redmon2015, pasquale2016iros, nguyen2016}).\\

\noindent{\bf Implementation Details.} We extensively experimented fine-tuning of {\em CaffeNet} and {\em GoogLeNet}, for which we report in the paper systematic and statistically robust performance trends on the proposed benchmark. Since {\em VGG-16} and {\em ResNet-50} have remarkably longer training times, for these models we performed less systematic experiments, that confirmed analogous trends (which we do not report).

After testing several hyper-parameters settings, we identified two fine-tuning ``regimes'' as representatives, which we selected for our analysis: one updating only fully-connected layers, while keeping convolution layers fixed, and another one more aggressively adapting all layers to the training set. We refer to these two protocols as {\em conservative} and {\em adaptive} and report their corresponding hyper-parameters in Tab.~\ref{tab:tuningStrategies} (the two are characterized by different learning rates for each network layer).

We refer to the supplementary material for our analysis of fine-tuning regimes and the model selection protocols implemented in our fine-tuning experiments.


\section{Results on Object Categorization}
\label{sec:experiments}

In this Section we present our empirical investigation of deep learning methods in the robotic setting of iCubWorld.

\begin{figure}[t]
\centering  
\includegraphics[width=\columnwidth]{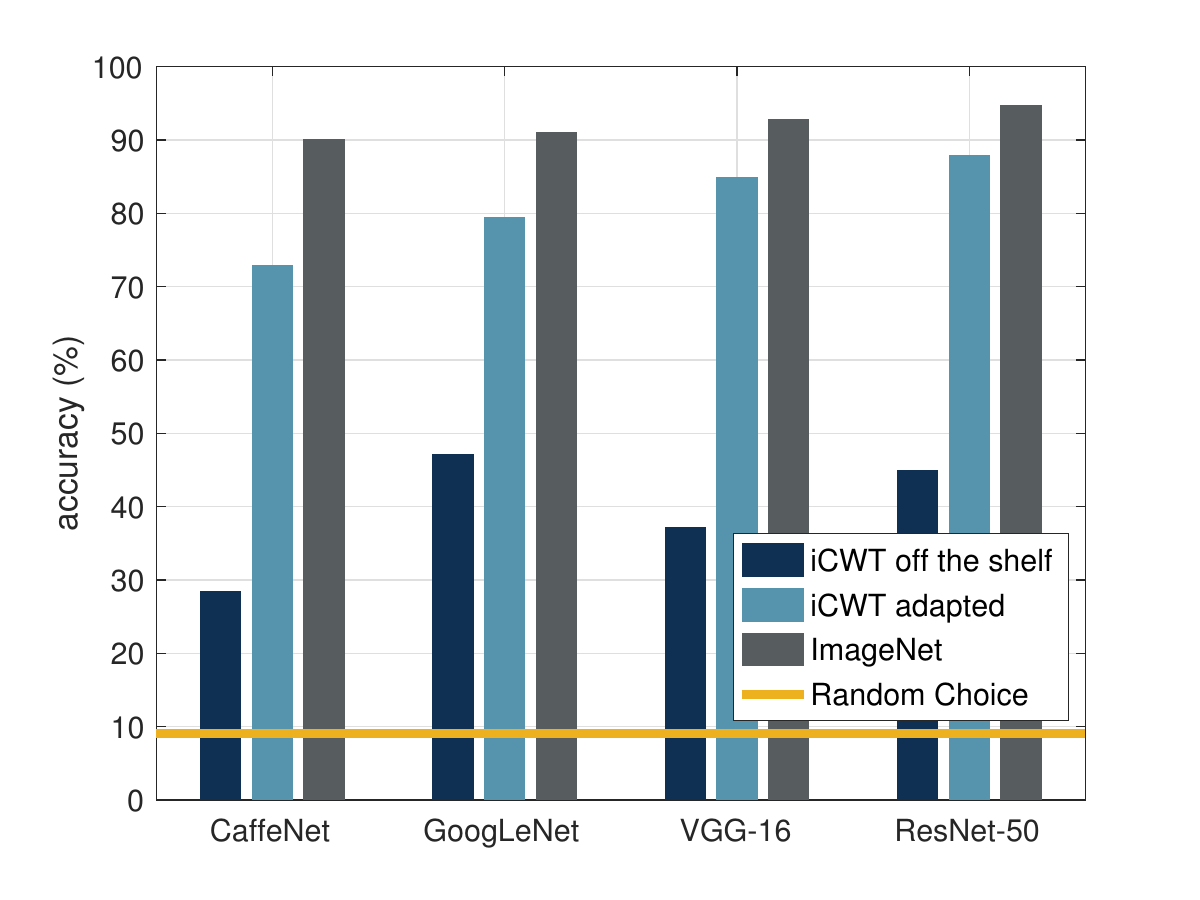}
\caption{Average classification accuracy of off-the-shelf networks (trained on ILSVRC) tested on iCWT (Dark Blue) or on ImageNet itself (Gray). The test sets for the two datasets are restricted to the $11$ shared categories (see Sec.~\ref{sec:icubworld}). (Light Blue) reports the classification accuracy when the same networks are ``transferred'' to iCubWorld (see Sec.~\ref{sec:ots} for details). The (Orange line) shows the recognition chance of a random classifier.}
\label{fig:off-the-shelf}
\end{figure}

\subsection{Deep Learning and (the Need for) Knowledge Transfer}
\label{sec:ots}

Modern datasets for visual recognition, as the ILSVRC, comprise million images depicting objects in a wide range of natural scenes. This extreme variability opens the question of whether datasets such as iCWT, which represent a smaller ``reality'', could be interpreted as their sub-domains. If this was the case, deep models trained on ImageNet would achieve high performance on iCWT as well.

To address this question, we evaluated four off-the-shelf CNNs (see Sec.~\ref{sec:deepCNNs}) for the task of image classification on iCWT. For this experiment we restricted the test set to the $11$ categories of iCWT that appear also in the ILSVRC (see Sec.~\ref{sec:icubworld} and Fig.~\ref{fig:20objects}). We compared these results with the accuracy achieved by the same models on the corresponding $11$ categories of the ImageNet dataset. The test set for iCWT was composed, for each category, by the images of all $10$ object instances, including all $5$ transformations for one day and the left camera (unless differently specified, we always used this camera for the experiments), for a total of $\sim 9000$ images per category. For testing on ImageNet, we downloaded the images of the corresponding $11$ categories (refer to the supplementary material for the synset IDs), comprising on average $\sim 1300$ images per category.

Fig.~\ref{fig:off-the-shelf} reports the average classification accuracy on iCWT (Dark Blue) and ImageNet (Gray). It can be observed that there is a substantial $\sim50-60\%$ performance drop when testing on iCWT. While a detailed and formal analysis of cross-domain generalization capabilities of deep learning models is outside the scope of this work (see, e.g., \cite{tommasi2015, hoffman2013,rodner2013}), in the supplementary material we qualitatively provide some interesting evidence of this effect. Note that in this case we have restricted the $1000$-dimensional output vector provided by the off-the-shelf CNNs, to the considered $11$ classes (we report in the supplementary material the same results when considering the entire $1000$-dimensional prediction).\\

\noindent{\bf Knowledge Transfer.} The performance in Fig.~\ref{fig:off-the-shelf} shows that all models performed much better than chance (Orange line), suggesting that the networks did retain some knowledge about the problem. Therefore, it seems convenient to transfer such knowledge, rather than training an architecture ``from scratch'', essentially by ``adapting'' the networks trained on ImageNet to the new setting (see Sec.~\ref{sec:transferlearning}).

In Fig.~\ref{fig:off-the-shelf} we report the classification performance achieved by models where knowledge transfer has been applied (Light Blue). For these experiments we followed the protocol described in Sec.~\ref{sec:feature-extraction}, where RLSC predictors are trained on features extracted from the deeper layers of the CNNs. We created a training set from iCWT by choosing $9$ instances for each category for training and keeping the $10$th instance for testing. We repeated the experiment for $10$ trials in order to allow each instance of a category to be used in the test set and Fig.~\ref{fig:off-the-shelf} reports the average accuracy over these trials. We observe a sharp improvement for all networks, which achieve a remarkable accuracy in the range of $\sim70-90\%$. While performance on ImageNet is still higher, such gap seems to be reduced for more recent architectures. What are the reasons for this gap? In the following we empirically address this question.

\subsection{Do we need more data?}
\label{sec:categorization}

While knowledge transfer can remarkably reduce the amount of data needed by deep CNNs in order to learn a new task, the size and richness of the training set remains a critical aspect also when performing transfer learning. A common practice to train deep CNNs in computer vision is to artificially augment the example images by applying synthetic transformations like rotations, reflections, crops, illumination changes, etc. From this perspective, in a robotic setting data augmentation can be achieved by simply acquiring more images (``frames'') depicting an object, while viewpoint or illumination change naturally. On the other hand, in a typical robotic application it is expensive to gather many different object instances to be shown to the robot.

In this Section, we investigate the impact of these aspects in robot vision, taking iCubWorld as a testbed. Note that in the following we use the term {\em instance} to refer to a specific object belonging to a given category, while {\em frame} denotes a single image depicting an object. 

\subsubsection{What do we gain by adding more frames?}
\label{sec:training-size-variability}

\begin{figure}[t]
\centering
\includegraphics[width=\columnwidth]{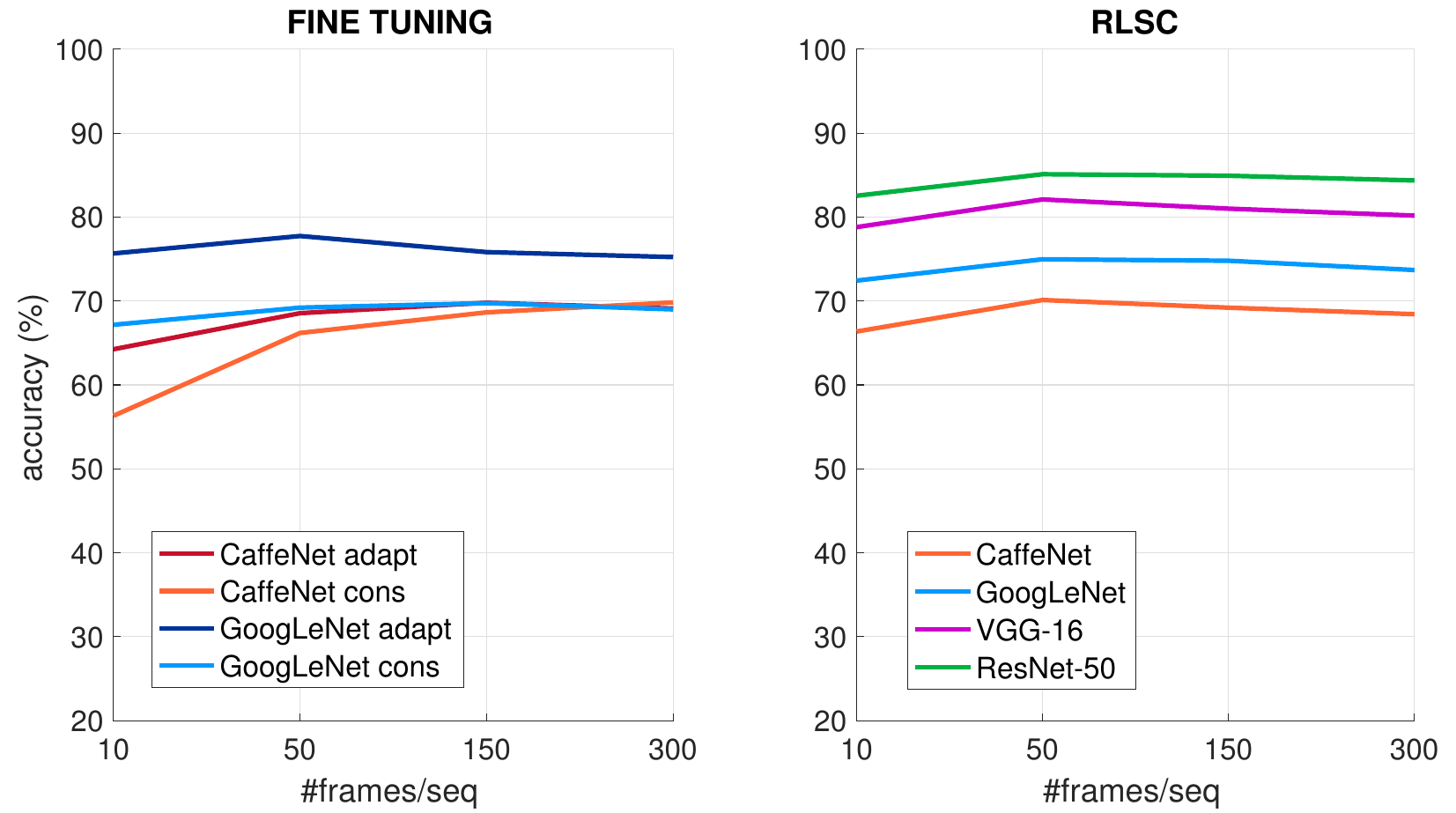}
\caption{{\bf Recognition accuracy vs \# frames}. (Left) Accuracy of {\it CaffeNet} and {\it GoogLeNet} models fine-tuned according to the {\em conservative} and {\em adaptive} strategies (see Sec.~\ref{sec:fine-tuning}). (Right) Accuracy of RLSC classifiers trained over features extracted from the $4$ architectures considered in this work (see Sec.~\ref{sec:feature-extraction}).}
\label{fig:training-size-variability}
\end{figure}

To compare the performance of models trained on an increasing number of frames, we consider a $15$-class categorization task on iCWT. For each category (see the supplementary material for their list) we used $7$ object instances for training, $2$ for validation and $1$ for testing.
We created training sets of increasing size by sampling randomly $N=10, 50, 150$ and finally $300$ frames from each image sequence of an object (we recall that each object in iCWT is represented by $10$ sequences containing $5$ isolated visual transformations acquired in $2$ days). 
Validation and test sets contained all images available for the corresponding instances. 
To account for statistical variability, we repeated the experiment for $10$ trials, each time leaving out a different instance for testing.
For this experiment, we considered only one of the two available days in iCWT. We used only the left camera, apart from when sampling $300$ frames, where we drew images also from the right camera. The number of frames per category therefore ranged from $350$ to $10500$.

Fig.~\ref{fig:training-size-variability} reports the average classification accuracy of different models as more example frames are provided. Surprisingly, most architectures achieve high accuracy already when trained on the smallest training set and show little or no improvement when new data is available. This finding is in contrast with our expectations, since increasing the dataset size does not seem key to a significant improvement in performance. To further support this finding, in the supplementary material we report results for the same experiment when using less example instances per category. Moreover, in Sec.~\ref{sec:rgbd-cat} we report similar observations on the Washington RGB-D dataset.

Secondary observations:
\begin{itemize}
\item[$\bullet$] {Fine-tuning and RLSC achieve comparable accuracy (both for {\it CaffeNet} and {\it GoogLeNet}).}
\item[$\bullet$] {We confirm the ILSVRC trends, with more recent networks generally outperforming older ones.}
\item[$\bullet$] {{\it CaffeNet} performs worse when training data is scarce because of the high number of parameters to be learned in the $3$ fully connected layers (see Sec.~\ref{sec:methods} and the supplementary material).}
\end{itemize}

\subsubsection{What do we gain by adding more instances?}
\label{sec:semantic-variability}

\begin{figure}[t]
\centering
\includegraphics[width=\columnwidth]{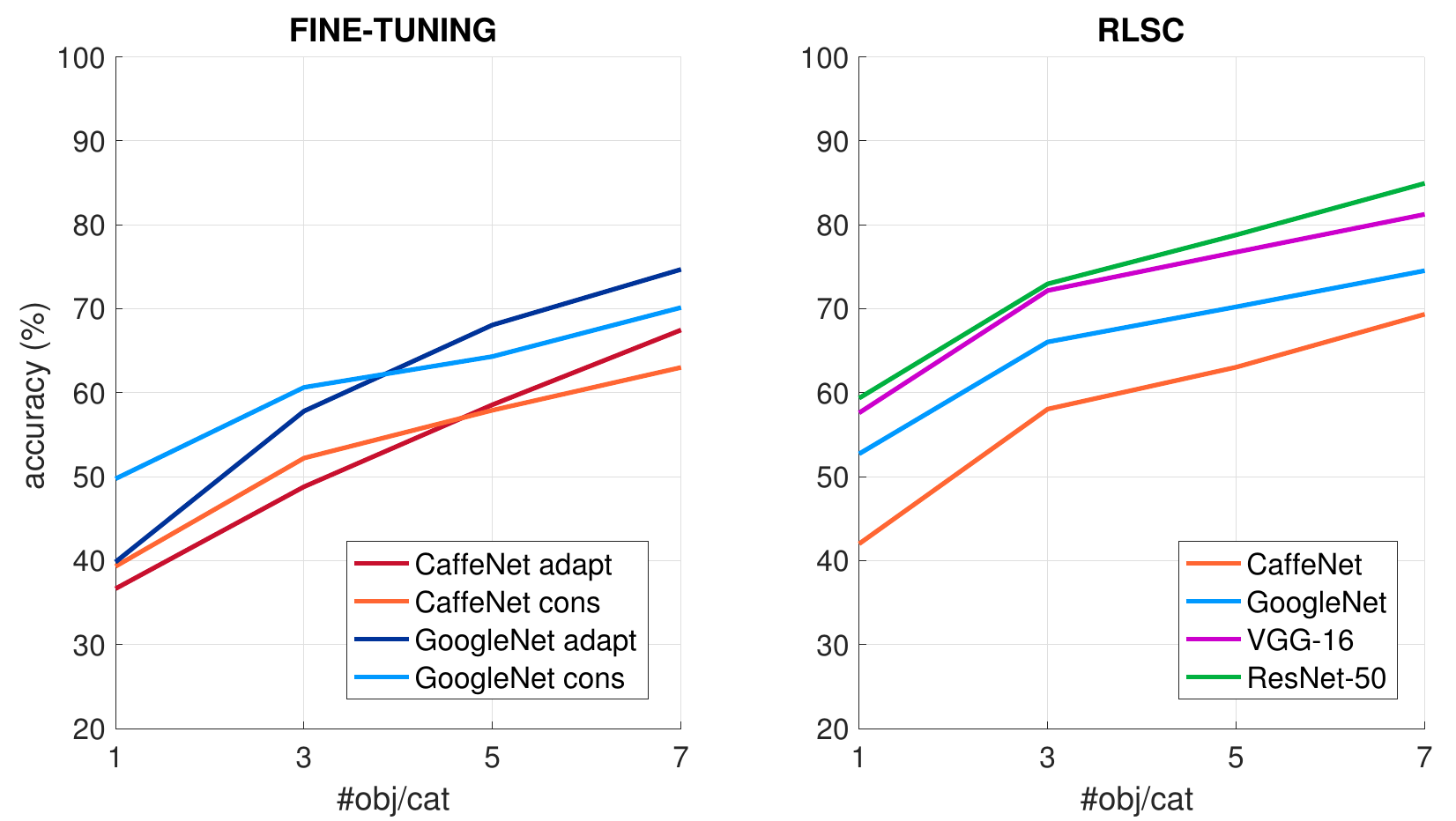}
\caption{{\bf Recognition accuracy vs \# instances} (number of object instances available during training). (Left) Accuracy of {\it CaffeNet} and {\it GoogLeNet} models fine-tuned according to the {\em conservative} and {\em adaptive} strategies (see Sec.~\ref{sec:fine-tuning}). (Right) Accuracy of RLSC classifiers trained over features extracted from the $4$ architectures considered in this work (see Sec.~\ref{sec:feature-extraction}).}
\label{fig:semantic-variability}
\end{figure}

We evaluated then the impact of experiencing less or more example object instances per category. We consider this process as increasing the {\em semantic} variability of the training set, in contrast to increasing the {\em geometric} variability by showing more frames of a given object.
To this end, we kept the same $15$-class categorization task of Sec.~\ref{sec:training-size-variability} and created multiple training sets each containing an equal number of $900$ frames per category, but sampled from an increasing number of instances per category, namely $1,3,5$ and $7$ (i.e., we took all $900$ frames from one instance per category, then $300$ frames from each of the $3$ instances per category, and so forth). Validation and test sets are as in the previous experiment and we repeated again the experiment for $10$ trials.

Results in Fig.~\ref{fig:semantic-variability} show that increasing the semantic variability dramatically improves the accuracy of {\it all} models by a similar margin (more than $20\%$), and performance does not saturate at $7$ example instances per category.
To further support this finding, in the supplementary material we report results for the same experiment when discriminating between even less categories ($10$ and $5$) and in Sec.~\ref{sec:rgbd-cat} we show a similar effect on the Washington RGB-D dataset.

Finally, it is worth noting that, when few example objects per category are available, {\em adaptive} fine-tuning (Dark Blue and Red) provides worst performance, suggesting that adapting ImageNet features (which are ``optimized'' for a rich categorization task) is not convenient if the dataset has poor semantic variability, as we also discuss in Sec.~\ref{sec:comparison}.

\begin{figure}[!t]
\centering
\includegraphics[width=\columnwidth]{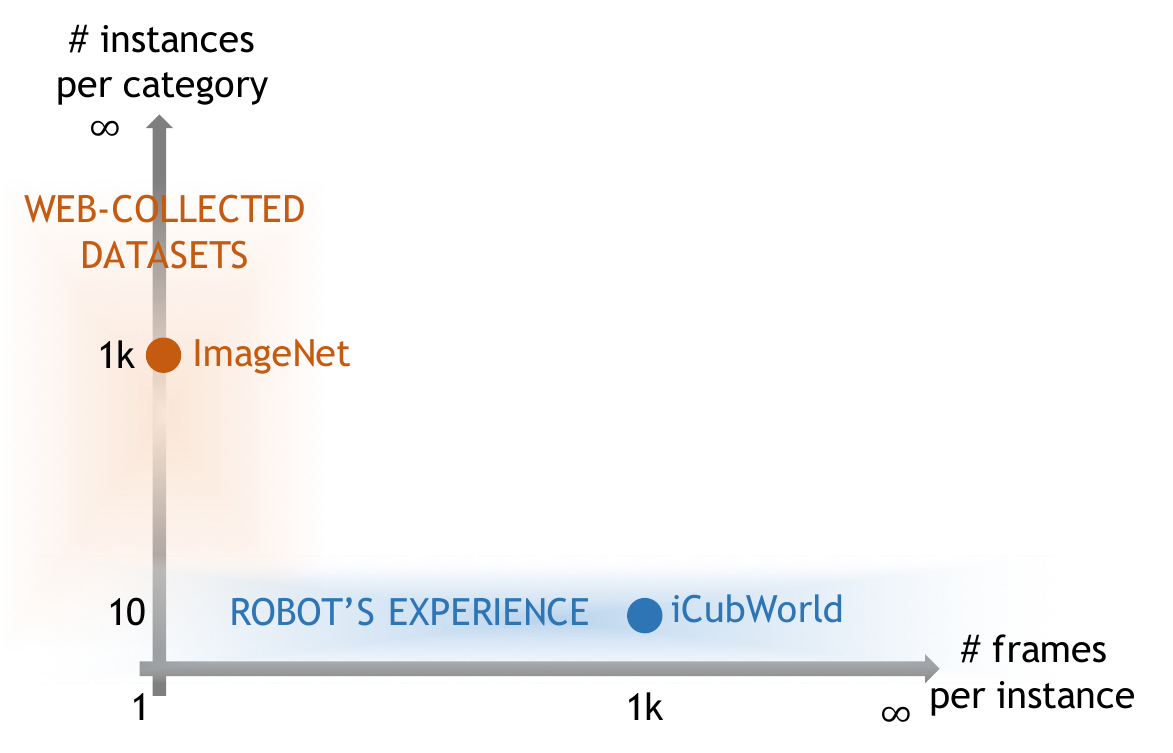}
\caption{Different training regimes in robot vision and image retrieval settings.}
\label{fig:robo-vs-img}
\end{figure}

\subsubsection{Robot Vision and Image Retrieval}
\label{sec:comparison}

In this Section we considered two major aspects associated to visual object recognition: semantic and geometric variability of objects' appearance. We observed that, while essentially the same problem is addressed both in robot vision and image retrieval, it is cast within two different regimes (see Fig.~\ref{fig:robo-vs-img}). In both cases it is possible to gather a large amount of image examples. However, typically, in image retrieval, each image depicts a different object instance (as an example, ILSVRC training set comprises $\sim1000$ images per category, $1$ image per instance). On the opposite end of the spectrum, in robot vision it is easy to gather many images of an object (the robot can observe it from multiple viewpoints), but there is a remarkable limitation in the number of instances that can be experienced.

\begin{figure*}[t]
\centering
\includegraphics[width=\textwidth]{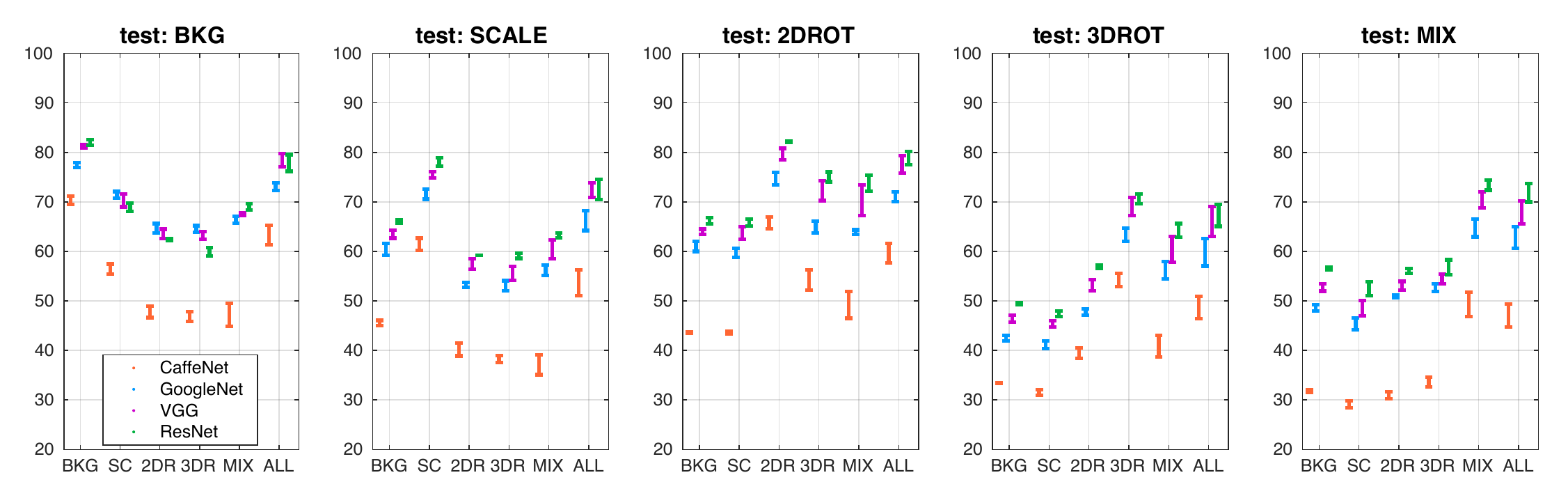}
\caption{{\bf Generalization performance across different visual transformations}. Models were trained on one of the $5$ transformations in iCWT (horizontal axis), and then tested on all transformations (title of the subplot). Accuracy is reported separately in each plot for each tested transformation. The bars indicate the (small) performance improvement achieved by including all frames of a sequence (higher end of bars) instead of subsampling $\sim 20$ of them (lower end of bars).}
\label{fig:invariance}
\end{figure*}

Our analysis has shown that the limited semantic variability that characterizes our setting dramatically reduces the recognition accuracy, even when relying on pre-trained deep learning models. Moreover, contrarily to our expectations, we observed that feeding more object views to the network (i.e., increasing the geometric variability) does not alleviate this lack: classifiers trained on very few, or many, object views, and fixed semantic variability, achieved identical performance.

This represents a problem, because usually a robot has access to limited instance examples of categories to be learned. In this perspective, adopting ``data augmentation'' strategies to artificially increase semantic variability, could be a viable solution. In Sec~\ref{sec:conclusions} we discuss how this problem may be addressed in the future, while in the following we focus our analysis on the invariance and robustness of deep representations to geometric (viewpoint) transformations.

\subsection{Invariance to Viewpoint Transformations}
\label{sec:invariance}
Invariance, i.e., robustness to identity-preserving visual transformations, is a desirable property of recognition systems, since it
increases the capability of generalizing the visual appearance of an object from a limited number of examples (ideally, just one)~\cite{anselmi2015, anselmi2016}.

In this Section, we investigate to what extent CNN models are invariant to the viewpoint transformations represented in iCWT.

We considered the same $15$-class categorization problem introduced in Sec.~\ref{sec:semantic-variability}, where we use $7$ instances per category for training, $2$ for validation and $1$ for testing. However, in this case we did not mix example images from all the $5$ available sequences ({\em $2$D Rotation, $3$D Rotation, Scale, Background} and {\em Mix}). Instead, we performed training (and validation) using only an individual transformation, and then tested the model on the others. We considered two different training set sizes: a larger one, with including all images from each sequence (i.e. $\sim 150\times7=1050$ images per category), and a smaller one, subsampling images of a factor of $7$ (i.e. $150$ images per category), in order to reproduce the two extreme sampling conditions as in Fig.~\ref{fig:training-size-variability}. 
As a reference, we considered a $6$-th training set ({\em All} ), of same size, obtained by randomly sampling images from the other $5$ training sets. This training set is analogous to those used in the experiments in Sec.~\ref{sec:semantic-variability} and~\ref{sec:training-size-variability}. We consider only one of the two available days in iCWT, since we aim to exclude nuisances due to illumination or setting changes that can happen from one day to another. We repeated the experiment for $10$ trials as in previous tests. 

Since in this Section we focus on the invariance of representations extracted from networks trained on ImageNet, we report only the accuracy of the approach based on training RLSC on off-the-shelf features (see Sec.~\ref{sec:feature-extraction}). We refer to~\cite{pasquale2017thesis} for additional results reported by fine-tuning strategies. Fig.~\ref{fig:invariance} shows the generalization capabilities of models trained on a single transformation (indicated on the horizontal axis) and tested on a different one (indicated in title of the subplot). Each bar starts at the accuracy achieved by the small training set and goes up to the accuracy achieved by the large training set, from which we suddenly note that there is no improvement by considering all frames from the sequence of a transformation, even on the transformation itself (i.e., all bars are very short). This explains and confirms the trend observed in Fig.~\ref{fig:training-size-variability}.

Overall, the classifiers perform well only on transformations that have been included -- even with a few examples -- in the training set: best performance is always achieved when training and test set include the same transformation, or when all transformations are included in the training set ({\it All}).
While generalization failure from $2$D to $3$D rotations of the object is expected, it is quite surprising that generalization also fails between affine transformations, namely {\it Scale}, {\it $2$D Rotation} and {\it Background}, which the CNN could have learned from ImageNet. We will investigate this aspect in Sec.~\ref{sec:invariance-id}, by studying the invariance of CNNs in the context of object {\em identification}.

It is finally worth noting that training on the {\it Mix} sequence achieves considerably good performance when tested on every specific transformation. This suggests that showing an object to the robot in a natural way, with transformations appearing in random combinations (instead of systematically collecting sequences comprising individual transformations), is a good approach to obtain predictors invariant to these transformations.


\section{Results on Object Identification}
\label{sec:identification}

Object identification is the task of discriminating between specific instances within a category, and it is clearly of paramount importance for robot to correctly interact with the environment.
In this Section we evaluate the performance of deep learning models on this task using iCubWorld.

The problem has been largely addressed with methods based on keypoints and template matching~\cite{lowe2004,philbin2008,collet2009,collet2011,collet2011b,muja2011,crowley2014}. However, it has recently been observed that approaches that rely on holistic visual representations perform typically better in scenarios characterized by substantial variations of objects' appearance~\cite{ciliberto2013}.
Following this, we focus the analysis on the methods considered in Sec.~\ref{sec:methods}.

\subsection{Knowledge Transfer: from Categorization to Identification }
\label{sec:training-size-variability-id}

We investigated knowledge transfer in the context of object identification, specifically to determine to what extent performance are affected by the number of frames per object. We addressed this question by building an experimental setting similar to the one investigated for categorization, i.e., considering tasks on iCWT with increasing number of example images. We selected $50$ objects from the dataset by taking all instances from the {\em book}, {\em flower}, {\em glass}, {\em hairbrush} and {\em hairclip} categories. Note that these categories do not appear in the ILSVRC. We created four training sets containing respectively $10,50,150$ and $300$ images per object, sampled randomly from the $4$ transformation sequences {\em $2D$ Rot, $3D$ Rot, Scale} and {\em Bkg} (see Sec.~\ref{sec:icubworld}). From each training set, $20\%$ images were retained for validation. The images from the {\em Mix} sequence were used to test the classification accuracy of the methods. As for the categorization experiment, only the images from a single day were used.

\begin{figure}[t]
\centering
\includegraphics[width=\columnwidth]{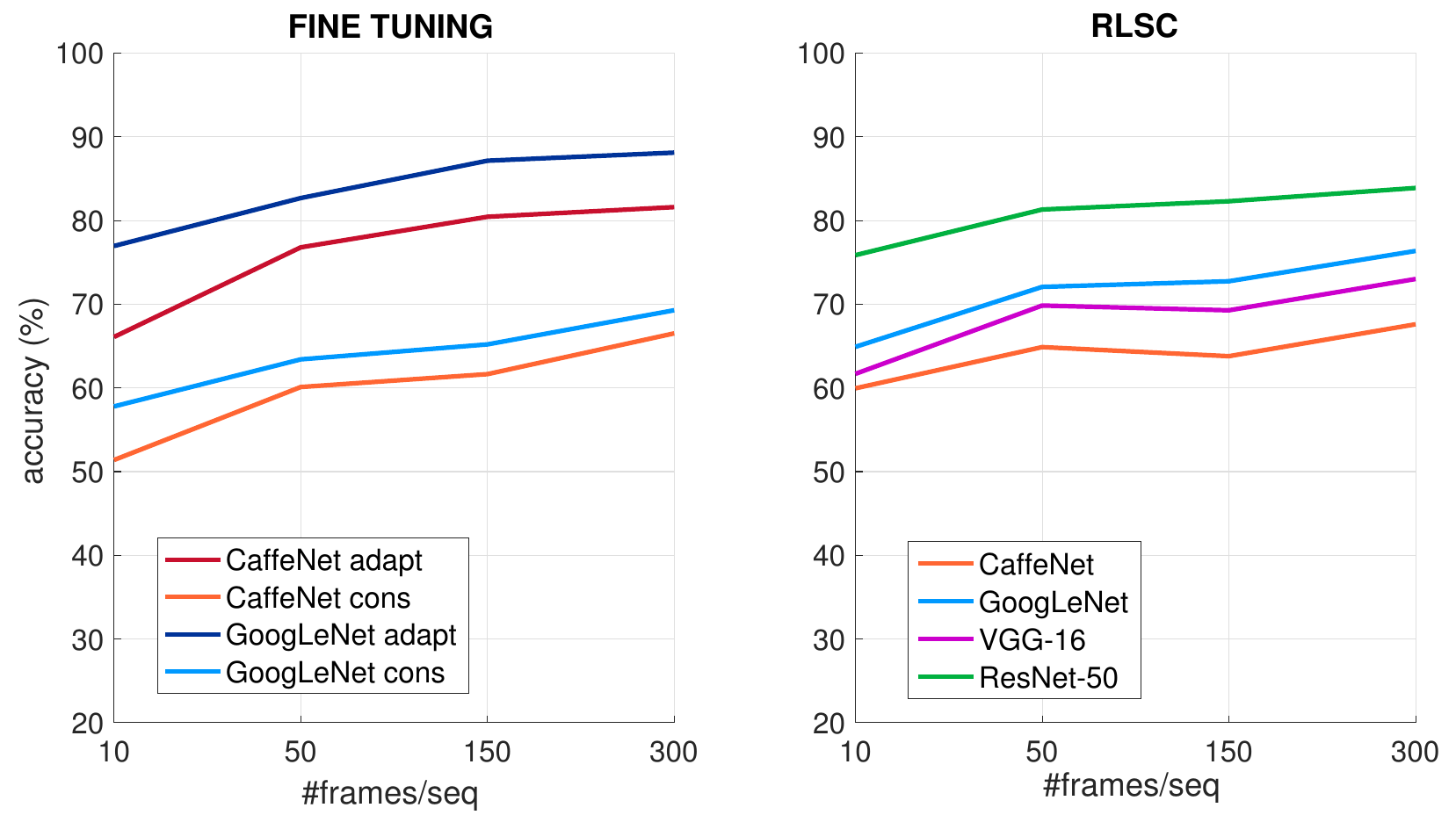}
\caption{{\bf Recognition accuracy vs \# frames (Identification)}. (Left) Accuracy of {\it CaffeNet} and {\it GoogleNet} models fine-tuned according to the {\em conservative} and {\em adaptive} strategies (see Sec.~\ref{sec:fine-tuning}). (Right) Accuracy of RLSC classifiers trained over features extracted from the $4$ architectures considered in this work (see Sec.~\ref{sec:feature-extraction}).}
\label{fig:training-size-variability-id}
\end{figure}

\begin{figure*}[t]
\centering
\includegraphics[width=\textwidth]{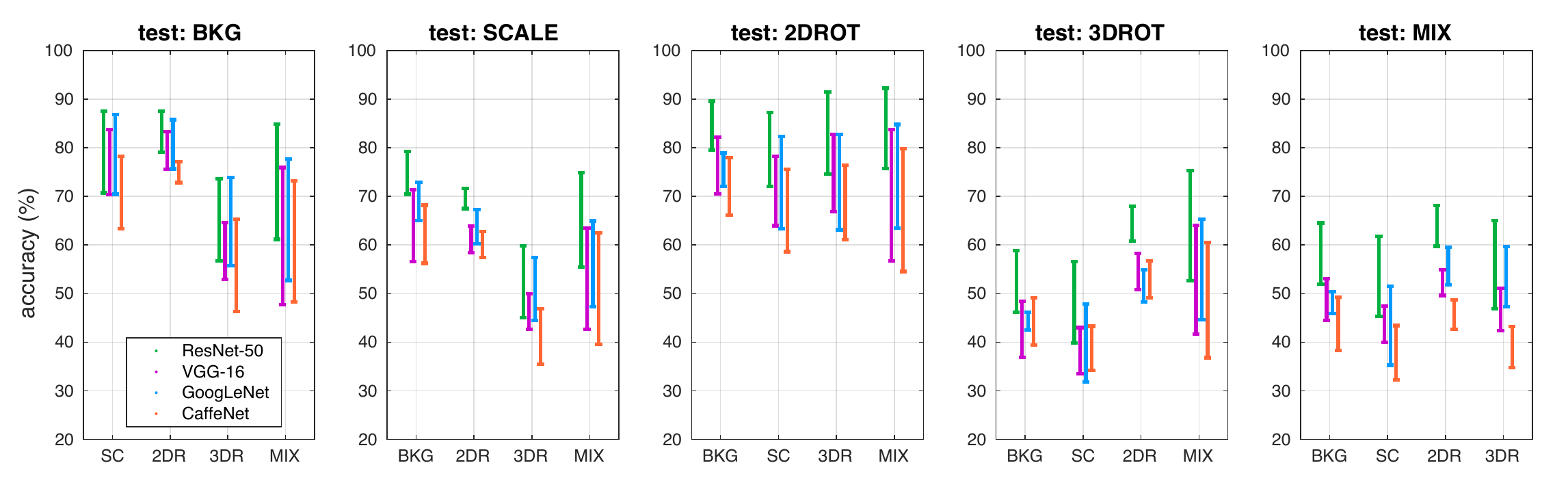}
\caption{{\bf Generalization performance across different visual transformations (Identification).} RLSC are trained on one of the $5$ transformations in iCWT (horizontal axis), and then tested on the other transformations (title of the subplot). Accuracy is reported separately in each plot for each tested transformation. The bars indicate the performance improvement achieved by including all frames of a sequence (higher end of bars) instead of subsampling $\sim 10$ of them (lower end of bars)}.
\label{fig:invariance-id}
\end{figure*}

Fig.~\ref{fig:training-size-variability-id} reports the average accuracy of models trained on a growing number of example images per object. Differently from what observed for categorization (Sec.~\ref{sec:training-size-variability}), in this case adding images of an object greatly improves performance (with a $15$-$20\%$ margin).
We also notice that, differently from categorization, in this case the {\em adaptive} fine-tuning (Dark Blue and Red) significantly outperforms the other strategies (RLSC and {\em conservative} fine-tuning).
This finding confirms recent empirical evidence~\cite{razavian2014} that adapting CNNs trained on ImageNet (rather than ``from scratch'') is beneficial also in identification tasks. It also confirms and extends results from the instance retrieval literature~\cite{babenko2014, gordo2016}, which show that several, different,  images of objects are necessary for such adaptation.

\subsection{Invariance to Viewpoint Transformations }
\label{sec:invariance-id}

Similarly to Sec.~\ref{sec:invariance},  we investigated object identification on isolated viewpoint transformations. To this end, we kept the same $50$-class identification task of Sec.~\ref{sec:training-size-variability-id} and restricted the training set to contain only images from a single transformation sequence among the $5$ available in iCWT. The resulting models were tested separately on the remaining transformations to assess their ability to generalize to unseen viewpoints. As in Sec.~\ref{sec:invariance} we considered one day and evaluated only methods based on training RLSC on top of off-the-shelf features.

Fig.~\ref{fig:invariance-id} reports the accuracy of models trained on a single transformation, using respectively $10$ frames per object (lower end of bars) or $150$ (upper end of bars), and tested on the others. It can be noticed that the improvement observed in Fig.~\ref{fig:training-size-variability-id} is confirmed and clarified in this setting. Adding images containing varied transformations (that is, from the {\it Mix} training set) helps to generalize to the individual transformations. In addition, adding images from a single transformation always improves performance on the others (also the profile of the upper ends of the bars is slightly flatter, i.e., with smaller gaps between one transformation to another).

We conclude that the viewpoint invariance of off-the-shelf representations is small, but adding more views of an object can remarkably boost it. While this is positive, in real world applications, where a robot typically has to learn novel objects on the fly, collecting extensive object views is probably unfeasible. To this end, in the following Section we report on a simple strategy we proposed to increase the invariance of CNNs and hence reduce the number of example frames needed to learn new objects ``online''.

\subsubsection{Improving Viewpoint Invariance of CNNs}
\label{sec:invariance-id-adapted}

The performance improvement achievd by the {\em adaptive} fine-tuning strategy (Fig.~\ref{fig:training-size-variability-id}) may be explained by the fact that adapting the inner layers of CNNs with images representing viewpoint variations improves the invariance of the internal network representation to such transformations.

In this Section we investigate this effect in detail. 
We consider the same learning setting of Fig.~\ref{fig:invariance-id}, i.e., RLSC trained on few example frames of a single visual transformation. Instead of off-the-shelf features, we compare using features from different {\it CaffeNet} or {\it GoogLeNet} models, previously fine-tuned on a selected image set from iCWT. Specifically, we consider the following fine-tuning strategies:

\begin{itemize}
\item[$\bullet$] {\bf iCubWorld identification (iCWT id)}. This set contains all transformation sequences available in iCWT for a number of objects. It is conceived to investigate if the CNN can learn to be invariant to experienced viewpoint changes. We used objects belonging to categories not involved later in the considered $50$-class identification task ({\em oven glove}, {\it squeezer}, {\em sprayer}, {\em body lotion} and {\em soda bottle}). Fine-tuning was performed with the {\em adaptive} strategy for an identification task. The validation set was obtained following the protocol of Sec.~\ref{sec:training-size-variability-id}.
\item[$\bullet$] {\bf iCubWorld categorization (iCWT cat)}. Same set as (iCWT id), but fine-tuning performed on a $5$-class categorization task. The validation set was obtained following the protocol of Sec.~\ref{sec:training-size-variability}. This set is conceived to test the same hypothesis of learning viewpoint invariance, but at the category rather than instance level. 
\item[$\bullet$] {\bf iCubWorld $+$ ImageNet (iCWT + ImNet)}. Same as (iCWT cat), but images of the $5$ categories are sampled also from ImageNet. Note that most iCWT categories do not appear in ILSVRC but are contained in synsets in the larger ImageNet dataset (see supplementary material for the synset list). This set is conceived to investigate if learning category-level viewpoint invariance jointly with semantic variability can be beneficial.
\item[$\bullet$] {\bf ImageNet (ImNet)}. This dataset differs from previous ones and contains the $5$ ImageNet synsets corresponding to the $5$ categories of the $50$ objects on which the CNN would be later used as feature extractor ({\em book}, {\em flower}, {\em glass}, {\em hairbrush} and {\em hairclip}, see Sec.~\ref{sec:training-size-variability-id}). Fine-tuning was performed on the categorization task. This set is conceived to investigate the effect of focusing the CNN on the categories of the objects to be discriminated later on, by learning category-level features from data available on-line.
\end{itemize}

\begin{figure*}[t]
\centering
\includegraphics[width=0.9\textwidth]{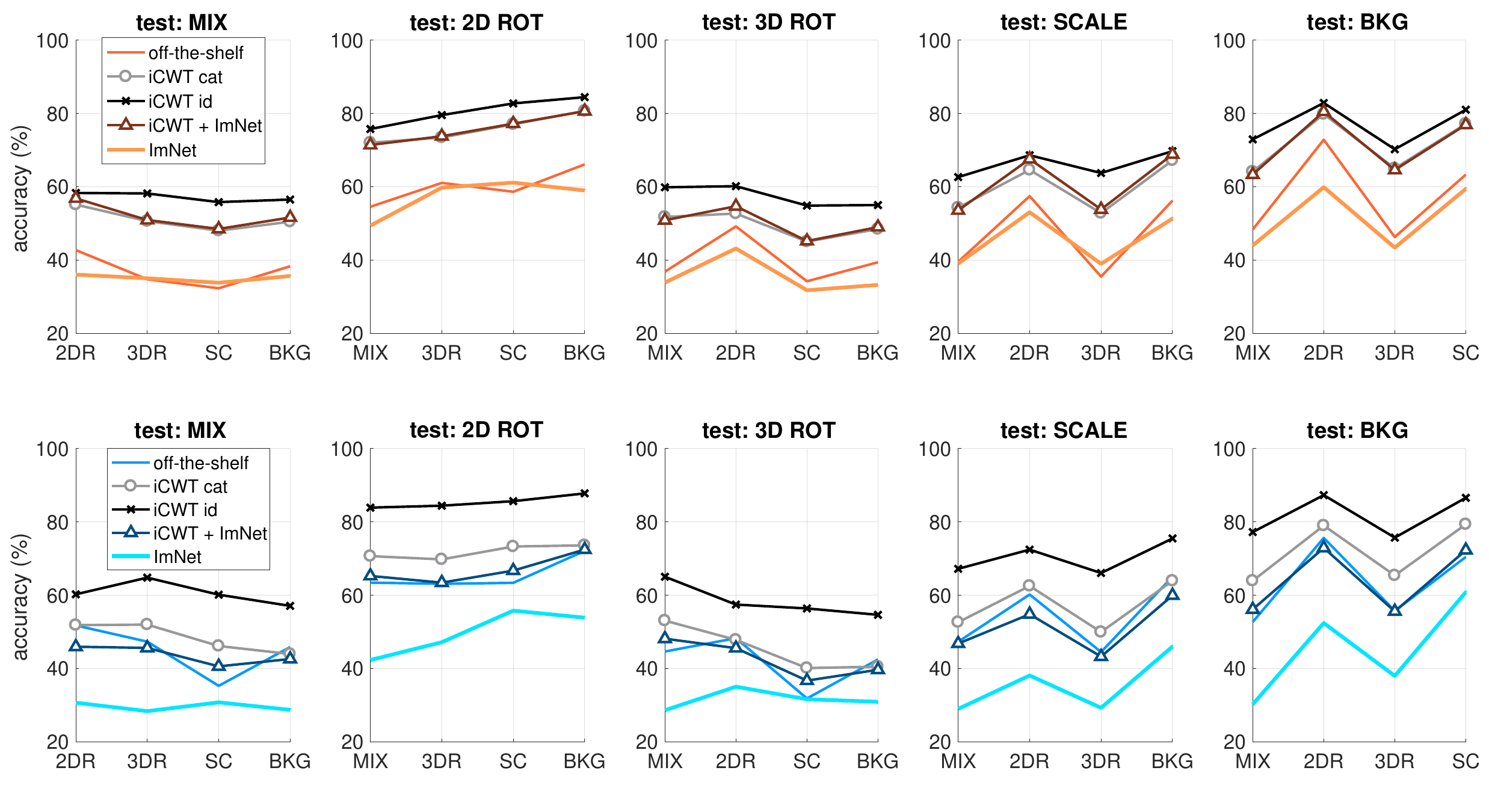}
\caption{Same experiment setting as in Fig.~\ref{fig:invariance-id}, but using different image representations, provided by {\it CaffeNet} (top, Orange) or {\it GoogLeNet} (bottom, Blue) network models, previously fine-tuned according to different strategies (see Sec.~\ref{sec:invariance-id-adapted}).}
\label{fig:invariance-id-adapt}
\end{figure*}

Fig.~\ref{fig:invariance-id-adapt} reports the accuracy of RLSC trained on features from {\it CaffeNet} (top, Orange) or {\it GooLeNet} (bottom, Blue) fine-tuned on each of these image sets. Training was performed identically to Sec.~\ref{sec:training-size-variability-id}, on the smaller set containing $10$ examples per instance. It can be noticed that preliminary fine-tuning on sequences available in iCWT for an identification task (iCWT id) is particularly effective, leading to the highest improvement over off-the-shelf features. We also observe that all approaches that involve preliminary fine-tuning over transformation sequences in iCWT do provide an accuracy increase. Interestingly, fine-tuning on corresponding categories in ImageNet (ImNet), degrades performance, possibly since it has increased invariance to intra-class variations between instances, which are in fact relevant for the identification task.

Comparing with Fig.~\ref{fig:invariance-id}, it can be observed that RLSC trained on (iCWT id) features from $10$ examples per object now performs better or on par with off-the-shelf features from $150$ examples per object. The proposed strategy is a viable approach to build robust feature extractors: training, possibly offline, a CNN to learn invariances by collecting images of objects undergoing various transformations and then using the resulting features online to train new classifiers using only few images.\\

\noindent{\bf Note on Object Categorization.} We repeated the experiment of Fig.~\ref{fig:invariance} using features from these pre-fine-tuned CNNs. However, our results (reported in supplementary material) showed that, in this setting, performance does not improve with respect to off-the-shelf features. This futrther confirms our findings in Sec.~\ref{sec:comparison}, that networks trained on ILSVRC are highly optimized for categorization and adapting them to datasets like iCWT does not provide a clear advantage.


\section{Results on Washington RGB-D}
\label{sec:rgbd}

In this Section we show that the conclusions drawn from our results on iCWT are supported by similar trends on other robotic datasets. Specifically, we consider Washington RGB-D (WRGB-D for short)~\cite{lai2011}.

WRGB-D is a turntable dataset comprising $300$ objects organized into $51$ categories. The number of objects per category ranges from $3$ to $14$ ($6$ on average). For each object, $3$ RGB-D sequences were acquired by a camera mounted at $\sim1$m from the object and $30, 45, 60^{\circ}$ elevation angles. In each sequence, the camera performed one revolution around the table at constant speed, recording $\sim250$ $640\times480$ frames. In the literature (and in our evaluation) a reduced dataset version is usually employed, including every fifth frame, cropped to tightly include the object, for a total of $41877$ images.\\

\subsection{Object Categorization Benchmark}
\label{sec:rgbd-cat}

The benchmark categorization task consists in discriminating between the $51$ categories, by leaving out one instance per category for testing. We considered this task in our experiments and, as in the literature, we averaged performance on the same $10$ trials released by the authors~\cite{lai2011}, each time leaving out different $51$ instances for testing. We used a random $20\%$ of the training set for validation. 

As a first sanity check, we evaluated our methods on this benchmark: Table~\ref{tab:rgbd-cat} compares our results with two recent works that achieve the state-of-the-art by employing {\it CaffeNet} with similar transfer learning approaches. In particular, in~\cite{schwarz2015} SVM classifiers are trained on off-the-shelf features, while in~\cite{eitel2015} the network is fine-tuned. Note that we consider results exploiting only RGB, without depth information. It can be observed that, with {\it CaffeNet}, our methods achieve comparable results, and the best performance is achieved by {\it ResNet-50} features. 

We then repeated this experiment by increasing either the number of (i) example instances per category or (ii) example images per object. The goal of this evaluation is to reproduce respectively the setting of Fig.~\ref{fig:semantic-variability} and Fig.~\ref{fig:training-size-variability}. 

When increasing the number of example instances per category, we fixed the overall number of example images such that, similarly to Sec.~\ref{sec:semantic-variability}, we first sampled all example images for a category from one instance (around $150$), then $50\%$ images from each of two instances, and so forth, until all instances from all categories were included. Since, differently from iCWT, in WRGB-D each category has a different number of available training instances (from $2$ to $13$ since one is left for testing), for each category we stopped adding instances when all were included. Specifically, while most categories have at least $3$ training instances available, we could train only $25$ categories (out of $51$) on $5$ instances, and just $10$ ($5$) categories on $7$ (or more) instances. Fig.~\ref{fig:rgbd}(a) reports results of this experiment.

Conversely, Fig.~\ref{fig:rgbd}(b-e) report results of experiments where we fixed the maximum number of example instances per category to different values ($1$, $2$, $3$ and $all$) and, for each value, we repeated the experiment multiple times by progressively increasing the number of sampled frames per instance (starting from $10$ to {\it all} available).

For this evaluation, we considered the three transfer learning techniques based on {\it CaffeNet}, reported in Fig.~\ref{fig:rgbd} in Orange (Dashed, RLSC; Bold, {\it conservative} fine-tuning) and Red (Bold, {\it adaptive fine-tuning}). We left out the other methods because we assessed how they compare both on iCWT and on the reference RGB-D categorization task.

These results replicate the findings of iCWT: adding new instances (slope in Fig.~\ref{fig:rgbd}(a)) leads to mich higher performance improvement than than adding example views (slopes in Fig.~\ref{fig:rgbd}(b-e)). In this latter case a ``jump'' in performance is achieved only when a new instance is added to the training set. Note that best results are achieved in Fig.~\ref{fig:rgbd}(a), when including all training instances, with just $\sim1/6$ of the frames of those usually used in the reference benchmark (which corresponds to the {\it all} training set in Fig.~\ref{fig:rgbd}(e)). We point out that performance saturates at $5$ instances per category in Fig.~\ref{fig:rgbd}(a) because, as explained above, in WRGB-D there are very few categories with more instances to be included.

\begin{table}[t]
\centering
\caption{{\bf Categorization benchmark on WRGB-D}: Comparison with state-of-the-art.}
\label{tab:rgbd-cat}
\begin{tabular}{r l}
\textbf{Method} & \textbf{Accuracy ($\%$)} \\
\toprule
Schwarz et al. 2015     & $83.1 \pm 2.0$ \\
Eitel et al. 2015           & $84.1$ $\pm$ 2.7 \\
\hline
\hline
CaffeNet RLSC                 & $83.1 \pm 2.8$\\
CaffeNet FT adapt           & $81.9 \pm 2.7$ \\
CaffeNet FT cons            & $83.6 \pm 2.4$\\
\hline

GoogLeNet RLSC              & $84.6 \pm 2.8$\\
GoogLeNet FT adapt      &  $82.4 \pm 2.8 $\\
GoogLeNet FT cons       & $83.9 \pm 2.3$\\
\hline
VGG-16 RLSC                  & $87.5 \pm 2.3$\\
\hline
{\bf ResNet-50 RLSC}           & {\bf  89.4 $\pm$ 3.1 }\\
\bottomrule                                   
\end{tabular}
\end{table}

\subsection{Object Identification Benchmark}
\label{sec:rgbd-id}

The identification task considered in the literature on WRGB-D~\cite{lai2011}, consists in discriminating between the $300$ objects in the dataset. For each object, the sequences recorded at $30^{\circ}$ and $60^{\circ}$ elevation angles are used for training, while the one recorded at $45^{\circ}$ for testing. We considered this task, using a random $20\%$ of the training set for validation.

We first evaluated our methods on this benchmark: Table~\ref{tab:rgbd-id} compares our results with the state-of-the-art achieved in~\cite{schwarz2015} and in~\cite{held2016}. This latter is recent paper that aims to improve performance on object identification by exploiting deep architectures (specifically, {\it CaffeNet}) previously fine-tuned on ``multi-view'' image sequences. Here we refer to the accuracy that they report by applying a fine-tuning protocol similar to our {\it conservative} strategy. 
From Table~\ref{tab:rgbd-id} it can be noticed that our pipeline provides better results, when relying on {\it CaffeNet}, and best performance is achieved again by {\it ResNet-50} features.

\begin{figure*}[t]
\centering
\includegraphics[width=0.95\textwidth]{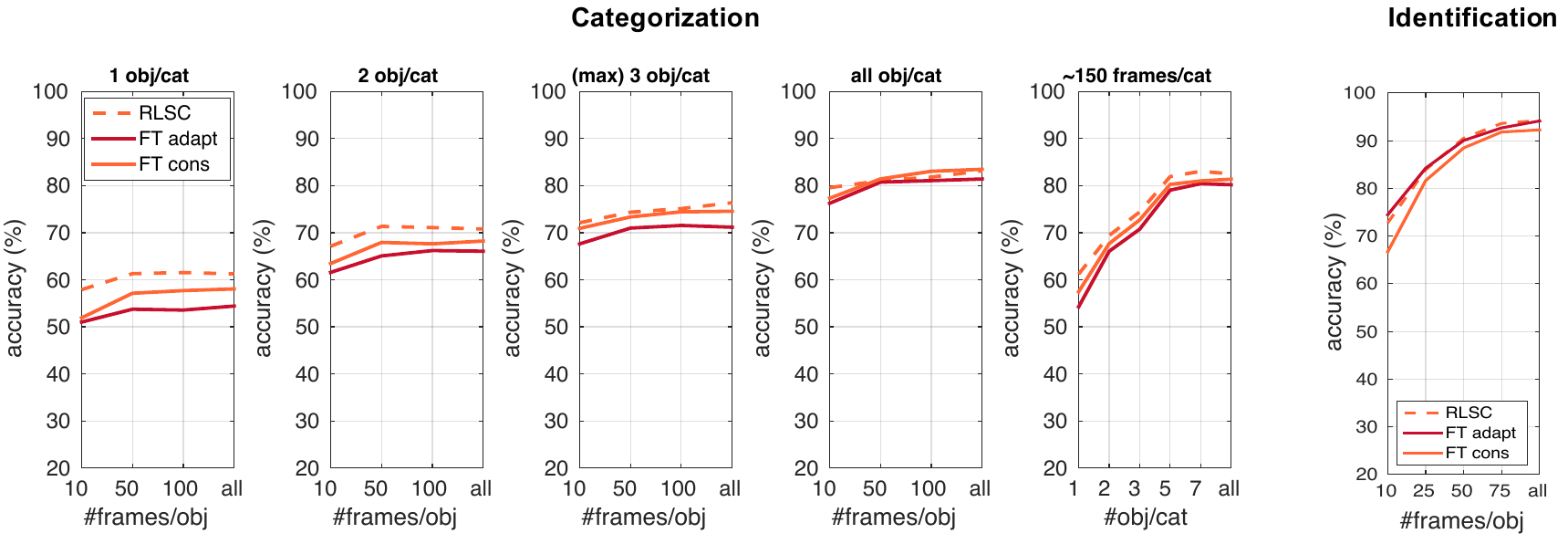}
\caption{{\bf Categorization and Identification on WRGB-D varying \# frames and \# instances.} (a) Object Categorization varying \# instances per category (horizontal axis). (b-e) Object Categorization varying \# frames per instance (horizontal axis), for different fixed numbers of instances per category (plot title). (f) Object Identification varying the number of \#frames per instance (horizontal axis). The three transfer learning methods indicated in the legend are applied to {\it CaffeNet}.}
\label{fig:rgbd}
\end{figure*}

Similarly to object categorization, we repeated the experiment by progressively increasing the number of example images per instance (from $10$ to {\it all} available). From the results reported in Fig.~\ref{fig:rgbd}(f), it can be noticed that, as for iCWT, and differently from the categorization setting, the performance gain provided on this task by adding object views to the training set is remarkable. In this case, the accuracy reported in the reference benchmark is achieved only when using all training images.

\begin{table}[t]
\centering
\caption{{\bf Identification benchmark on WRGB-D}: Comparison with state-of-the-art.}
\label{tab:rgbd-id}
\begin{tabular}{r l}
\textbf{Method} & \textbf{Accuracy ($\%$)} \\
\toprule
Schwarz et al. 2015     & $92.0$ \\
Held et al. 2016            & $93.3$ \\
\hline
\hline
CaffeNet RLSC                 & $94.0$  \\
CaffeNet FT adapt           & $94.0$ \\
CaffeNet FT cons            & $92.7$\\
\hline
GoogLeNet RLSC              & $94.3$ \\
GoogLeNet FT adapt      &  $93.9$ \\
GoogLeNet FT cons       & $92.5$ \\
\hline
VGG-16 RLSC                  & $94.5$ \\
\hline
{\bf ResNet-50 RLSC  }            & {\bf 96.0} \\
\bottomrule                       
\end{tabular}
\end{table}

\section{Discussion and Future Work}
\label{sec:discussion}

In this work we reported on an extensive experimental validation of the application of latest deep learning methods to visual object recognition in robotics. We challenged these methods on a setting that was specifically designed to represent a prototypical real world scenario. Our results showed that deep learning leads to remarkable performance for both category and instance recognition. We showed also that proper adoption of knowledge transfer strategies -- in particular mixing deep learning with ``shallow'' classifiers -- plays a key role, in that they leverage on the visual representation learned on large-scale datasets to achieve high performance in the robotic application.

However, a substantial gap still exists between the performance that can be obtained in the two domains. Our analysis shows that one reason for this gap is limited semantic variability of data collected in robotics settings (due to the intrinsic cost of acquiring training examples). Moreover, we need to push further these requirements since the error rate of robotic recognition systems will need to be as close as possible to zero in order to be considered for production and deployment in real applications. 

In this Section, we consider directions for future research to address these limitations and present how, in our opinion, the visual recognition problem could be addressed in robotics.\\

\noindent{\bf Improving Invariance}. Our experiments in Sec.~\ref{sec:invariance} and~\ref{sec:invariance-id} showed that off-the-shelf CNNs are mostly {\em locally} invariant, i.e., their representation is robust to small viewpoint changes, but these models can learn specific invariances from image sequences that are easy to collect in robotics settings. While local invariance has been of main interest to computer vision in the past~\cite{lowe2004,mikolajczyk2004}, current research on invariance is focused on learning representations robust to more ``global'' visual transformations~\cite{anselmi2015,anselmi2016}. 
Our results confirm that improving invariance is crucial to perform visual recognition in the real world~\cite{pinto2008,pinto2011,borji2016,poggio2016}. The role of viewpoint invariance in object categorization, which evidenced unexpected results in our setting, must also be further investigated (to this end see also~\cite{zhao2016,zhao2016b}).\\

\noindent{\bf ``Augmenting'' Semantic Variability: Real and Synthetic Data}. Our results pointed out how this aspect is key to object categorization. The simplest method for increasing semantic variability is to share data acquired from different robot platforms. 
This strategy has been used to learn hand-eye coordination on a manipulator~\cite{levine2016b}. Similarly, the goal of the Million Object Challenge~\cite{oberlin2015} is to create a sharing platform for data acquired by laboratories owning a Baxter robot\footnote{\url{http://www.rethinkrobotics.com/}}. Along a similar direction, we plan to extend iCubWorld with the help of the community of the iCub robot (more than $30$ research groups worldwide). This could also allow to extend the analysis presented in this work by introducing much larger variability. Indeed, this is another critical aspect that we started to address in the supplementary material.

A complementary approach follows the idea of data augmentation. More or less sophisticated synthetic image transformations (e.g., $2$D rotation, flip, crop/scaling, background and illumination changes) are already standard practice to simulate variations of instance appearance. Visual augmentation to cope for semantic variability is a more challenging problem, although recent work on inverse graphics~\cite{mansinghka2013,kulkarni2015} is a starting point in this direction. The potentials of this approach are limitless since, by synthetic generation, objects and environment parameters (viewpoint, lighting, texture, etc.) can be endlessly created and tuned to the application requirements. Generalization to real conditions would then clearly depend on the realism of simulated data, but the domain shift could be eventually tackled with transfer learning approaches. The work of~\cite{Handa2016CVPR,zhang2016physically} for indoor scene recognition and ShapeNet datasets~\cite{shapenet2015,Wu2015CVPR} datasets are examples of this strategy.\\

\noindent{\bf Integrating $3$D Information}. The integration of depth and RGB information with deep CNNs has been recently proved to help object recognition in robotics~\cite{schwarz2015,eitel2015,redmon2015,carlucci2016}.
The spatial structure of the scene can also be exploited as an additional self-supervisory signal, or prior information, to make the prediction more robust. Examples are the work of~\cite{song2015}, where memory about the room helps discriminating between old and novel objects. Interestingly,~\cite{pillai2015} implement a SLAM-aware detection system where the object's $3$D position is projected back to frames in order to help the recognition. With this approach, object instances could also be autonomously discovered by the robot.\\

\noindent{\bf Exploiting Temporal Coherence}. While it is important to push the limits of visual recognition at the image level, a robot is typically exposed to a continuous stream of frames, where visual information is correlated. Accuracy may be remarkably improved by exploiting this correlation, ranging from simple solutions as temporal averaging of predictions~\cite{pasquale2015} to more complex architectures including recent recurrent networks~\cite{donahue2015}.
The temporal correlation among consecutive frames can be exploited as self-supervisory signal, too. Recent works exploit this information to learn visual representations in absence of frame-level annotations~\cite{wang2015,goroshin2015,goroshin2015b,jayaraman2015,agrawal2015}.\\

\noindent{\bf Self-supervised Learning}. We opted for the help of a ``teacher'' for the acquisition of iCWT because we needed a relatively fine control on the object movements to isolate viewpoint transformations. However, making the acquisition self-supervised, 
i.e., implementing explorative strategies through which the robot autonomously interacts with the environment to collect examples, would allow to extend iCubWorld datasets, while limiting human effort.
Training instances in this scenario could be gathered autonomously by detecting invariances in the data which correspond to physical entities (e.g., coherent motion patterns~\cite{wang2015} or bottom-up saliency cues).
Strategies specific to the robotic domain could be devised by integrating multiple sensory modalities~\cite{sinapov2014,higy2016} and a repertoire of explorative actions~\cite{montesano2008,fitzpatrick2003,hogman2016,pinto2016}).\\

\noindent{\bf Multi-task Learning}. 
In this work we adopted a standard classification approach where no similarity or relation is assumed among classes. However, objects and categories have common features (think to object parts like wheels, legs, and so forth). Incorporating information about class similarities can significantly improve performance, especially if training data is limited.
In the literature on multi-task learning several approaches have been proposed to enforce these relations on the learning problem, when they are available a-priori~\cite{evgeniou2005,joachims2009,fergus2010}, or to learn them, when unknown~\cite{argyriou2008,minh2011,dinuzzo2011,ciliberto2015b}.\\

\noindent{\bf Incremental Learning}. 
A robotic recognition system should guarantee reliability in lifelong scenarios, by learning to adapt to changing conditions.
In line with the literature on ``learning to learn'' and transfer learning~\cite{thrun95,tommasi2010,kuzborskij2013}, strategies to exploit knowledge of previous, well-represented classes, in order to improve prediction accuracy on novel, under-represented ones, ultimately would be a key component of robotic recognition systems to be deployed in real world applications, as observed in recent work~\cite{camoriano2017,sun2016}.


\section{Conclusions}
\label{sec:conclusions}

In this paper we presented a systematic experimental study on the application of deep learning methods for object recognition to robot vision settings. For our tests we have devised a prototypical vision task for a humanoid robot in which human-robot interaction is exploited to obtain realistic supervision and train an object recognition system. We presented the iCWT dataset and an in-depth investigation of the performance of state-of-the-art CNNs applied to our scenario. Results confirm deep learning is a remarkable step forward. However, there is still a lot that needs to be done to reach the level of robustness and reliability required by real world applications. We identified specific challenges and possible directions of research to bridge this gap. We are confident that the next few years will be rich of exciting progress in robotics.

\begin{acknowledgements}
The work described in this paper is supported by the Center for Brains, Minds and Machines (CBMM), funded by NSF STC award CCF-1231216; and by FIRB project RBFR12M3AC, funded by the Italian Ministry of Education, University and Research. We gratefully acknowledge NVIDIA Corporation for the donation of a Tesla k40 GPU used for this research.
\end{acknowledgements}


\bibliographystyle{spbasic}      
\bibliography{biblio_thesis_giulia}   

\ 
\tabularnewline
\newpage


\appendix
\twocolumn[\section*{\LARGE Are We Done with Object Recognition? The iCub Robot's Perspective. [Supplementary Material]} \section*{\large\normalfont Giulia Pasquale, Carlo Ciliberto, Francesca Odone, Lorenzo Rosasco and Lorenzo Natale}\section*{ }]



\section{Objects in the iCWT Dataset}

\begin{figure*}
\centering
\includegraphics[width=0.85\textwidth]{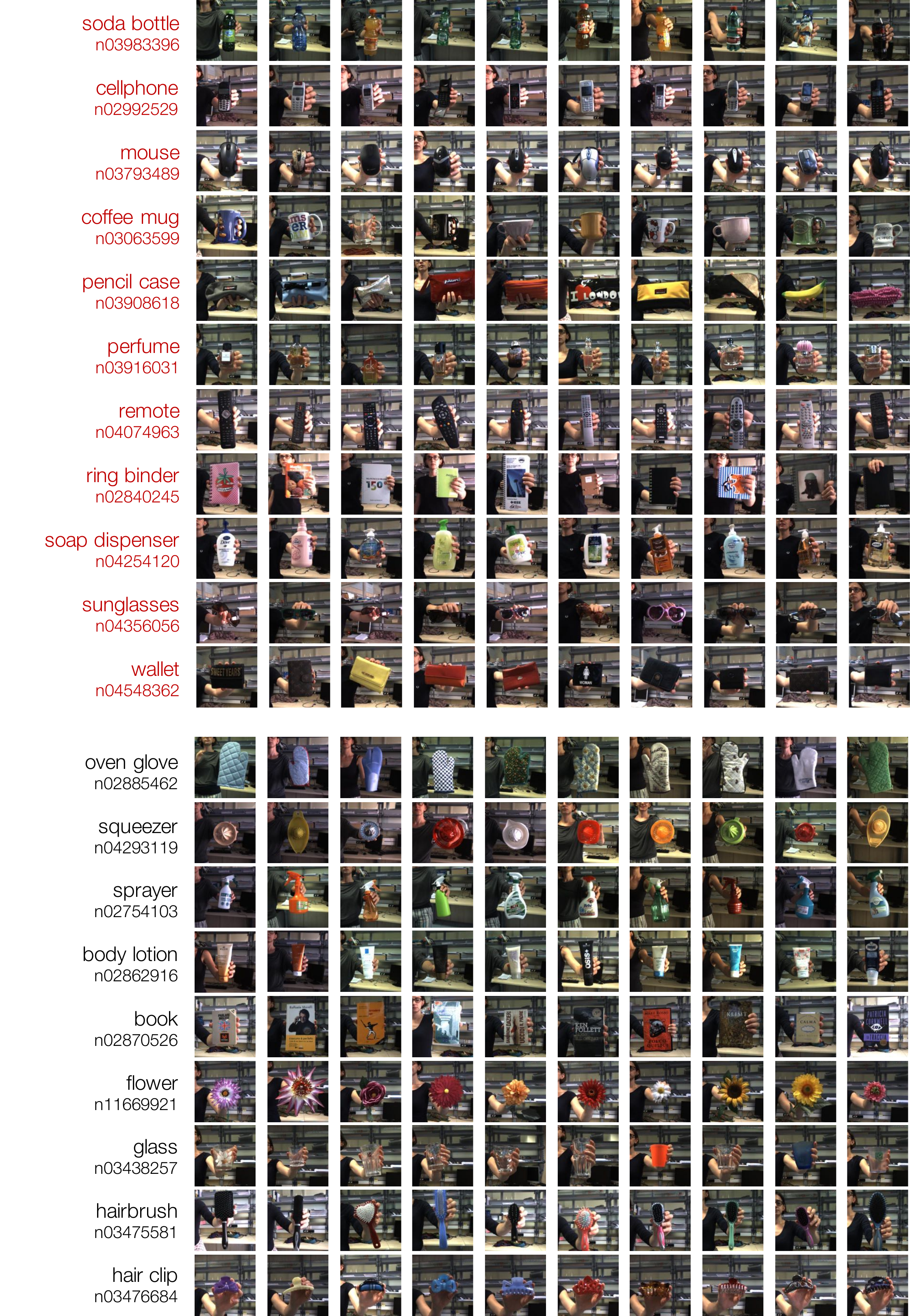}
\caption{Example images for the $200$ objects in iCubWorld Transformations.}
\label{fig:200objects}
\end{figure*}

Fig.~\ref{fig:200objects} shows one example image for each object in the iCubWorld Transformations (iCWT) dataset. We report the associated ImageNet synset in Red for categories among the $1000$ classes of the ILSVRC~\cite{russakovsky2015}, in Black for others~\cite{deng2009}.

\section{Testing Off-the-shelf CNNs}

In Fig.~\ref{fig:off-the-shelf} we observed that testing off-the-shelf CNNs to iCWT leads to poor accuracy. In this Section we provide details on this experiment and some qualitative explanations for the observed performance drop. 

\subsection{Image Preprocessing}
\label{sec:preproc}

We first evaluated the impact of processing with CNNs coarser or finer regions around the object of interest. Specifically, we compared extracting from the images either (i) a square region of fixed radius centered on the object or (ii) the bounding box provided by depth segmentation (obtained as explained in Sec.~\ref{sec:acquisitionsetup}).
In both cases, we included more or less background (respectively fixing the radius to $256$ or $384$ and leaving a margin of $30$ or $60$ pixels around the bounding box). 

\begin{figure}[t]
\centering  
\includegraphics[width=\columnwidth]{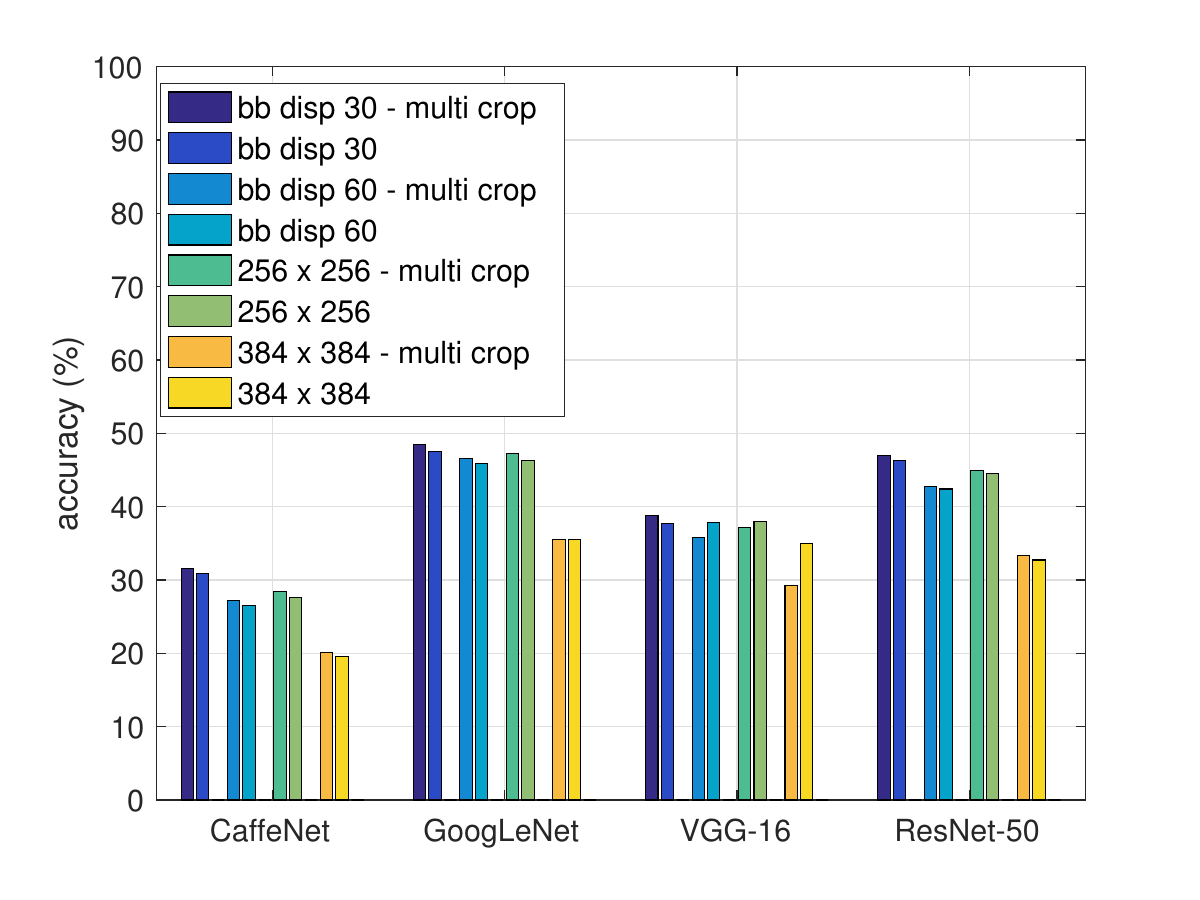}
\caption{Average classification accuracy of off-the-shelf networks tested on iCWT segmenting the object according to different strategies. }
\label{fig:crops}
\end{figure}

Then we reproduced the operations described at the reference page of {\sc Caffe} models (see Tab.~\ref{tab:preproc}). These basically consist in subtracting the mean image (or the mean pixel) of the training set and running the CNN on a grid of fixed-size crops ($227 \times 227$ for {\it CaffeNet} and $224 \times 224$ for the other models) at multiple scales. The final prediction is computed by aggregating the predictions of the crops. Since, however, in our case, the considered region is already cropped around the object, we also tried considering only one central crop. 
Fig.~\ref{fig:crops} reports the experiment of Fig.~\ref{fig:off-the-shelf}, when testing off-the-shelf CNNs on iCWT with the described pre-processing options. It can be noticed that (i) a finer localization of the object (Blue and Green) provides better performance and (ii) considering more than the central crop provides a little or no advantage at all. 

Based on this finding, for all the experiments reported in the paper we opted for extracting a square $256 \times 256$ region from the image (whose resolution is $640\times480$) and considering only the central crop (Light Green in Fig.~\ref{fig:crops}). 
We opted for a fixed-size region rather than the bounding box from depth segmentation, since this latter has varying shape and should be resized anysotropically to be processed by the CNNs, thus impacting on the viewpoint transformations we aim to study.
Note that when applying the networks to ImageNet, we used the multi-crop strategy suggested for each architecture.

\begin{table}[t]
\centering
\caption{Image preprocessing executed before feeding the networks.}
\label{tab:preproc}
\begin{adjustbox}{max width=\columnwidth}
\begin{tabular}{l c c c}
\multirow{2}{*}{ {\bf Model}} & {\bf Mean } & \multirow{2}{*}{ {\bf Scaling}} & {\bf Crop} \\ 
& {\bf Subtraction} & & {\bf Extraction} \\
\toprule
{\bf CaffeNet}      & \multirow{2}{*}{image} &  mean image size & $2 \times 2$ grid + center \\
{\bf ResNet-50}						  &                                      & ($256 \times 256$) & mirrored \\
\midrule
\multirow{2}{*}{{\bf GoogLeNet}} &  \multirow{2}{*}{pixel}   &   \multirow{2}{*}{$256 \times 256$} & $2 \times 2$ grid + center \\
					  &                                      & & mirrored \\
\midrule
\multirow{3}{*}{{\bf VGG-16}}    & \multirow{3}{*}{pixel} & shorter side to   & $5 \times 5$ grid  \\
& & $256$, $384$, $512$ & at each scale \\
& &  & mirrored \\
\bottomrule                                      
\end{tabular}
\end{adjustbox}
\end{table}

\begin{figure}[t]
\centering  
\includegraphics[width=\columnwidth]{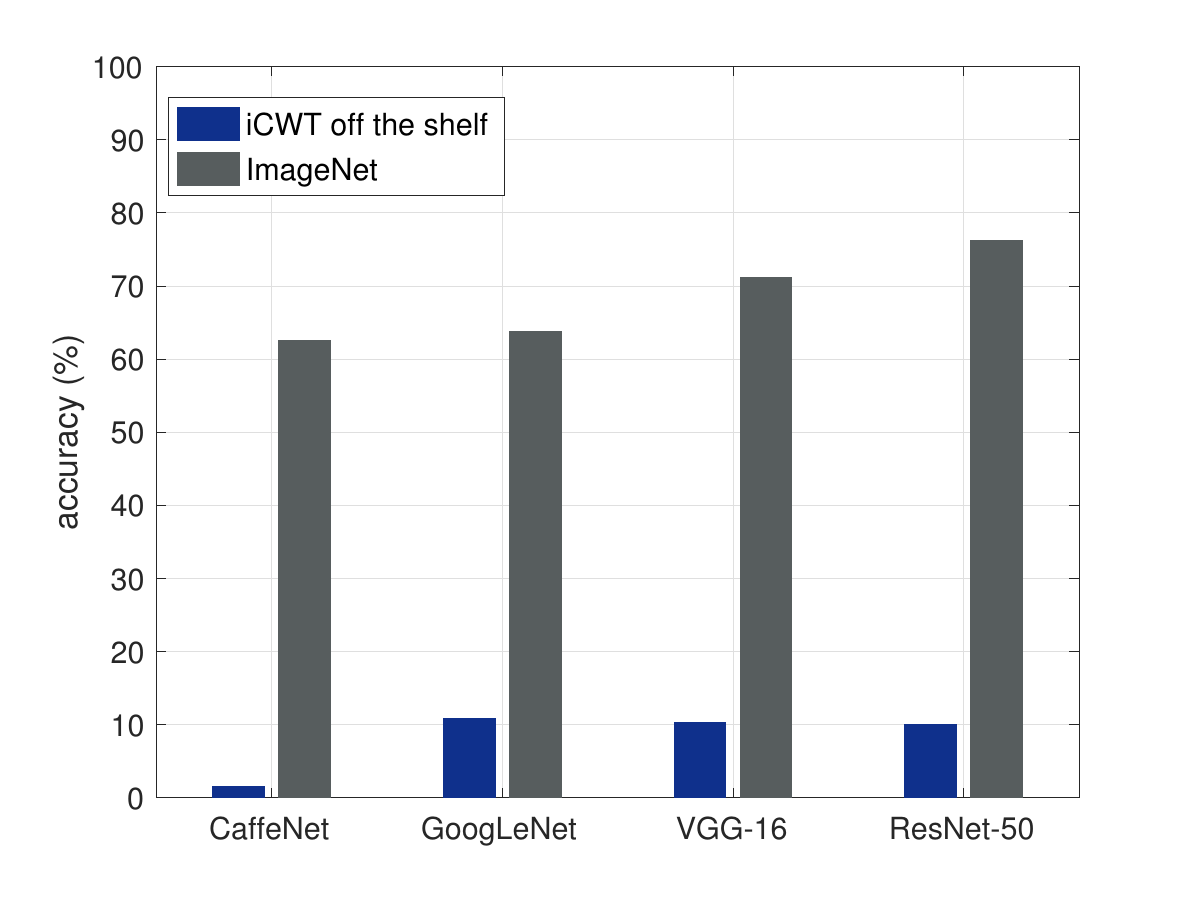}
\caption{Average classification accuracy of off-the-shelf networks (trained on ILSVRC) tested on iCWT (Dark Blue) or on ImageNet itself (Gray). The test sets for the two datasets are restricted to the $11$ categories in common (see Sec.~\ref{sec:icubworld}).}
\label{fig:ots1000}
\end{figure}

\begin{figure}[ht]
\centering
\includegraphics[width=0.8\linewidth]{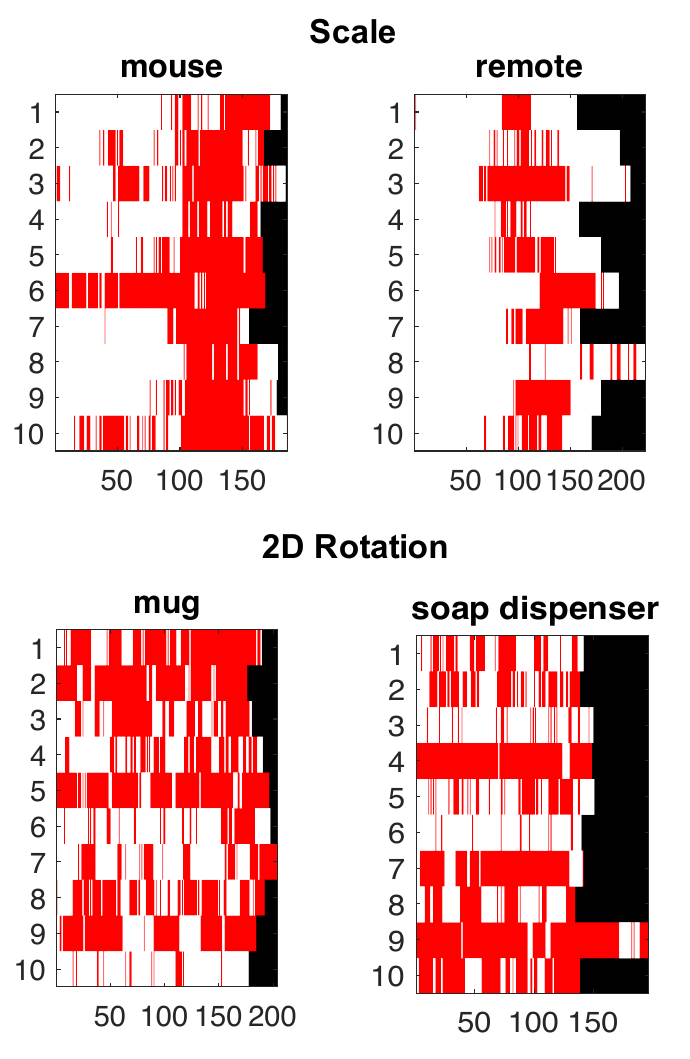}
\caption{Frame-level predictions of {\it GoogLeNet} on image sequences containing the {\it Scale} (Top) and {\it $2$D Rotation}  (Bottom) transformations, reported for $2$ representative categories. The sequences of the $10$ instances belonging to each category are represented as matrix rows (frame index in the horizontal axis). In each row, frame predictions over the sequence are represented as vertical bars: $White$ if correct, $Red$ if wrong ($Black$ if the sequence is ended). }
\label{fig:pred01-scale}
\end{figure}

\subsection{$1000$-class Categorization Results}

In Fig.~\ref{fig:off-the-shelf} we compared the accuracy of off-the-shelf CNNs tested on iCWT and ImageNet, selecting the $11$ scores of the considered classes from the vector of $1000$ scores produced by the CNNs trained on the ILSVRC. In Fig.~\ref{fig:ots1000} we report the accuracy of the same experiment, when considering $1000$ scores. As it can be noticed, performance drops significantly and proportionally for all models and both test sets (since now chance level for the model is $1/1000$). Note that, in this way, the models provide worse-than-chance predictions on iCWT, if chance level for iCWT is considered $1/11$).

\subsection{Viewpoint Biases}

In this Section we follow-up Sec.~\ref{sec:ots} and show potential biases in ImageNet, that may prevent off-the-shelf CNNs to generalize to iCWT. Keeping our observations qualitative, we analyze frame-level predictions in order to understand which views are actually ``harder'' to recognize and compare them with prototypical examples in ImageNet.

We report results for {\it GoogLeNet} ({\it CaffeNet} behaved similarly) and a subset of the test set considered in Fig.~\ref{fig:off-the-shelf}: specifically, $2$ representative categories and {\it Scale} and {\it $2$D Rotation} image sequences. As in Fig.~\ref{fig:off-the-shelf}, the prediction was computed as the maximum among the $11$ scores selected from the CNN output. In Fig.~\ref{fig:pred01-scale}we report frame-level predictions on {\it Scale} (Top) and {\it $2$D Rotation} (Bottom) sequences, separately per category. In each plot, rows represent the sequences of the $10$ instances of the category: the frame index is reported in the horizontal axis and each frame is a vertical bar: $White$ if the prediction is correct, $Red$ if it is wrong, $Black$ if the sequence is finished. 

It can be noted that, since during the {\it Scale} acquisition the operator was moving the object back and forth in front of the robot (starting close and moving backward), the CNN fails when the object is far, hence at a smaller scale (red bars mostly concentrated in the right half of rows). This confirms recent studies~\cite{herranz2016} pointing out that the scale bias of object-centric datasets as ImageNet prevents CNNs trained on them to generalize to real world, scene-centric, settings. Another example regards {\it $2$D Rotation} sequences, where the operator was rotating the object, keeping the same face visible, at a constant speed. In these sequences it can be noticed that the CNN fails at periodic time intervals, corresponding to less common views in ImageNet. In fact, soap dispensers and mugs in ImageNet are mostly placed on tables and, consequently, the CNNs fails when these are rotated from the vertical.


\section{Model Selection}

In this Section we provide details on the hyper-parameter selection for the methods adopted in the paper (Sec.~\ref{sec:transferlearning}).

\subsection{Feature Extraction and RLSC}
\label{sec:paramsel-nystrom}

We illustrate some design choices for the pipeline presented in Sec.~\ref{sec:feature-extraction}. 

To extract CNN representations, we considered fully-connected layers providing a vector global image representation.
These were specified for each architecture in Tab.~\ref{tab:feat_extraction}: in particular, for {\it CaffeNet} and {\it VGG-16} we tried either {\it fc6} or {\it fc7} layers as explained in the following.

We then resorted to~\citep{rudi2015} to implement RLSC with Gaussian Kernel and Nystr\"{o}m subsampling. The major hyper-parameters of this algorithm the number $m$ of training examples sampled to approximate the kernel matrix, the Gaussian's $\sigma$ and the regularization parameter $\lambda$. While the latter two were assigned based on standard cross-validation, that we performed for each experiment, in the following we report how we empirically determined a reasonable range for $m$.

\begin{figure}[!t]
\centering
\includegraphics[width=\columnwidth]{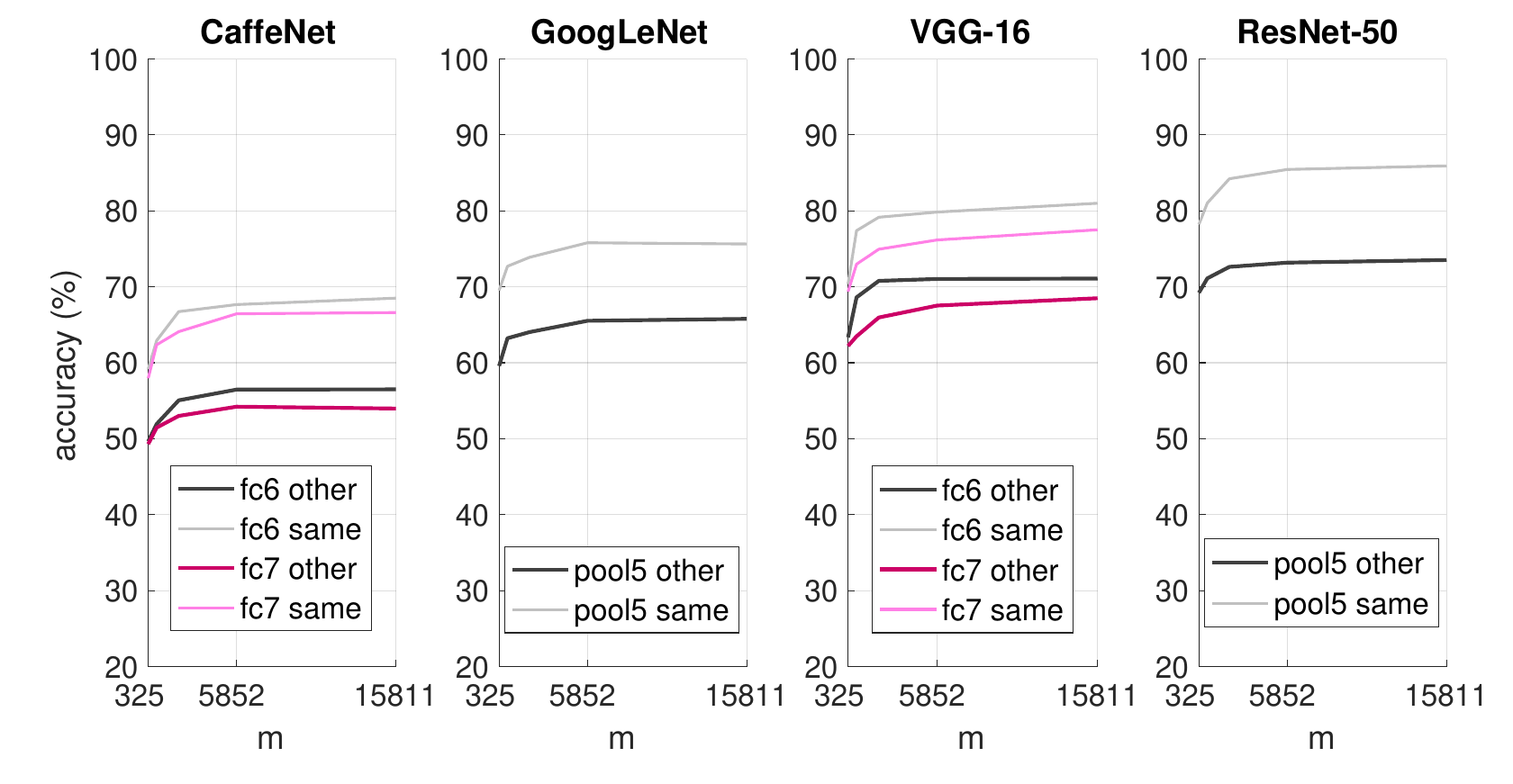}
\caption{Accuracy reported by training RLSC on image representations extracted by the four considered CNNs. A ``large'' categorization experiment is performed ($\sim N=95$K training examples) and $m$ is varied between $\sqrt N $ and $15$K (horizontal axis). Performance on the same day of training (Light colors) and on a different one (Dark colors) is reported.}
\label{fig:nystrom-full}
\end{figure}

We considered two categorization tasks on iCWT, representative of the smallest and larger tasks that we expected to run in our analysis: respectively $\sim 170$ and $\sim 6300$ examples per class, for $15$ classes (similarly to Sec.~\ref{sec:training-size-variability}). 
Fig.~\ref{fig:nystrom-same} and~\ref{fig:nystrom-full} report the average accuracy respectively for the small and large experiment. We used image representations from the four considered architectures, using either {\it fc6} layer (Gray) or {\it fc7} (Pink) for {\it CaffeNet} and {\it VGG-16}. We increased logarithmically the value of $m$ (horizontal axis) starting from $\sqrt N$, $N$ being the size of the training set. For the small experiment we stopped at $N$, whereas for the large experiment we stopped at values where we observed very little performance improvement. Performance on the same day of training is reported in light Gray/Pink, on a different day in dark Gray/Pink.

From this experiment we observed that {\it fc6} features consistently outperformed {\it fc7}, hence we decided to use this layer when extracting representations from off-the-shelf {\it CaffeNet} or {\it VGG-16}. Then, we observed that relatively small values of $m$ could provide good accuracies with far smaller training times and therefore we selected $ m = \min( $15$K ,  N ) $ in all experiments.

It is worth noting that we performed a similar experiment (not reported) with CNNs previously fine-tuned on subsets of iCWT and, in that case, {\it fc7} features were better. Hence, we used this layer in the experiments of Sec.~\ref{sec:invariance-id-adapted}. This is in line with recent work showing that, while lower layers provide more ``general'' features, more ``specialized'' features can be extracted from higher layers in models trained on closer domains.

\subsection{Fine-tuning}

In this Section we provide details on the fine-tuning procedures and the definition of the two {\it adaptive} and {\it conservative} strategies reported in Tab.~\ref{tab:tuningStrategies}.

\subsubsection{General Protocol}

\begin{figure}[ht]
\centering
\includegraphics[width=\columnwidth]{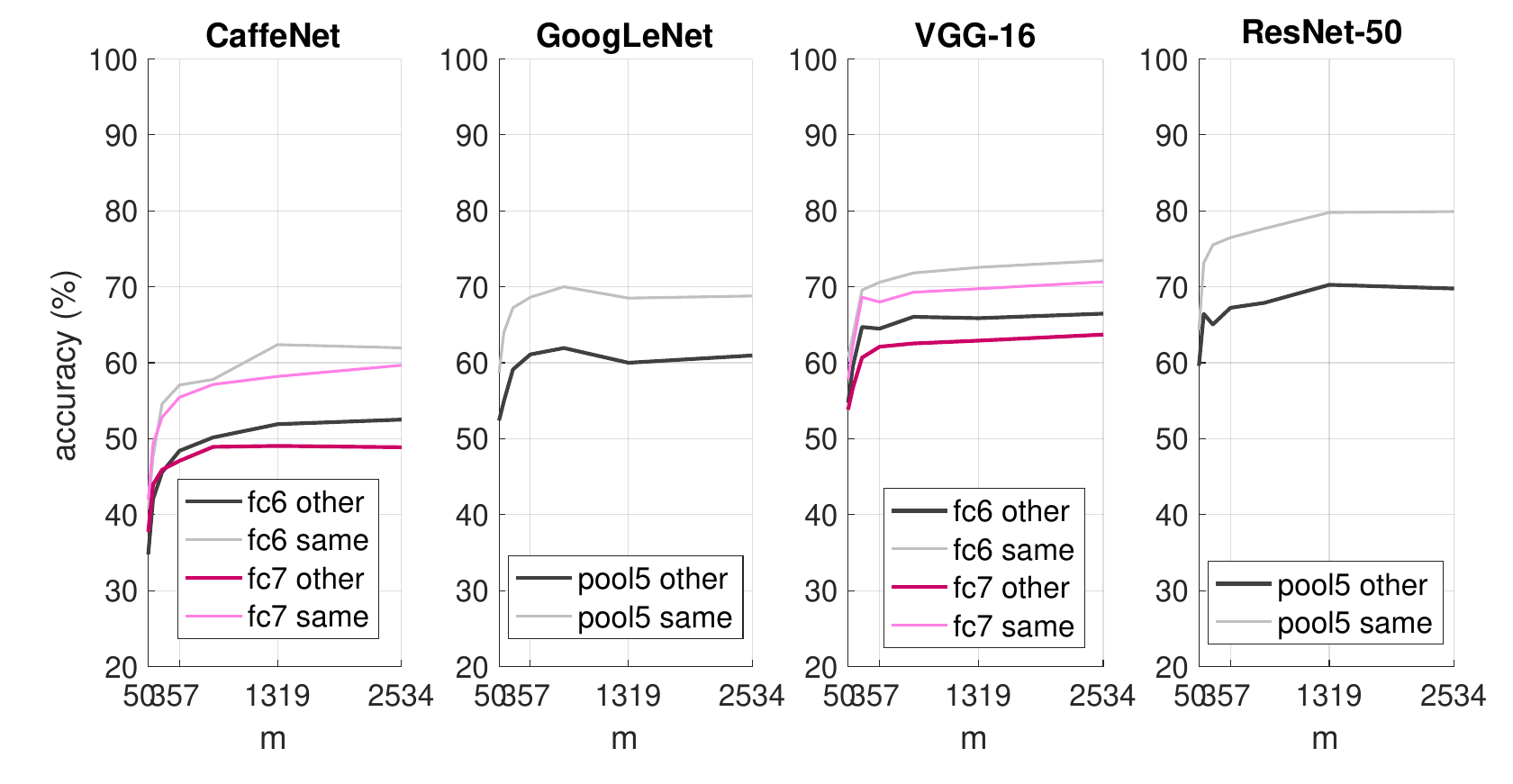}
\caption{Similar to Fig.~\ref{fig:nystrom-full} but on a ``small'' categorization experiment ($\sim N=2500$ training examples), varying $m$ between $\sqrt N $ and $N$.}
\label{fig:nystrom-same}
\end{figure}

Image preprocessing was similar to the one described in~\ref{sec:preproc} (except that, as per {\sc Caffe} standard, during training a random $227 \times 227$ crop, randomly mirrored, was extracted).
Importantly, the training set was shuffled, since we observed that similarity of images within a mini-batch (as consecutive frames would be) negatively affected convergence. We evaluated performance on a validation set every epoch and finally chose the model at the best epoch.
We set the mini-batch size to values specified in {\sc Caffe} (reported in Tab.~\ref{tab:tuningStrategies}).
The number of epochs was fixed empirically, observing that performance saturated after $\sim6$ epochs but for the {\it conservative} fine-tuning of {\it CaffeNet}, which involves learning the parameters of the $3$ fully-connected layers and takes more epochs to converge (we stopped at $36$).

\subsubsection{Hyper-parameters Choice}

While model selection is per se an open problem when dealing with deep networks, in our setting this issue is even more complicated by the fact that we do not target a fixed reference task, but plan to span over a wide range of tasks comprising small and large training sets. To this end in this Section we report on the empirical analysis we set up in order to understand the effect and relative importance of the (many) hyper-parameters involved in fine-tuning deep architectures as {\it CaffeNet} or {\it GoogLeNet}.

We considered the same two ``small'' and ``large'' categorization tasks adopted to perform parameter selection for RLSC as described in~\ref{sec:paramsel-nystrom} and fine-tuned {\it CaffeNet} and {\it GoogLeNet} by varying the values of multiple hyper-parameters. In the following, we report this analysis separately for the two architectures.\\

{\bf CaffeNet.} We considered the parameters in Tab.~\ref{tab:tuningStrategies} and varied them in the following way:

\begin{itemize}
\item {\tt Base LR}: the starting learning rate of the layers that are initialized with the parameters of the off-the-shelf model. We tried $10^{-3}$, $5*10^{-4}$, $10^{-4}$, $5*10^{-5}$, $10^{-5}$, $10^{-6}$, $0$.
\item {\tt Learned FC Layers}: which fully-connected (FC) layers are learned from scratch with their specific starting LR. We tried to learn (i) only $fc8$ with starting LR set to $10^{-2}$, or (ii) including also $fc7$ and (iii) finally also $fc6$. As an empirical rule, every time we included one more layer to learn from scratch, we decreased the starting LR of these layers of a factor of $10$ (hence $10^{-3}$ in (ii) and $10^{-4}$ in (iii)).
\item {\tt Dropout}: percentage of dropout in FC layers. We tried $50\%$ (default {\sc Caffe} value) or $65\%$.
\item {\tt Solver}: the algorithm used for the stochastic gradient descent. We used the {\it SGD} solver~\cite{bottou2012} in {\sc Caffe}.
\item {\tt LR Decay Policy}: the decay rate of the learning rates. We tried either polynomial decay with exponent $0.5$ or $-3$, or step decay decreasing the LR of a factor of 10 every $2$ epochs.
\end{itemize}

\begin{figure}[!t]
\centering
\includegraphics[width=\columnwidth]{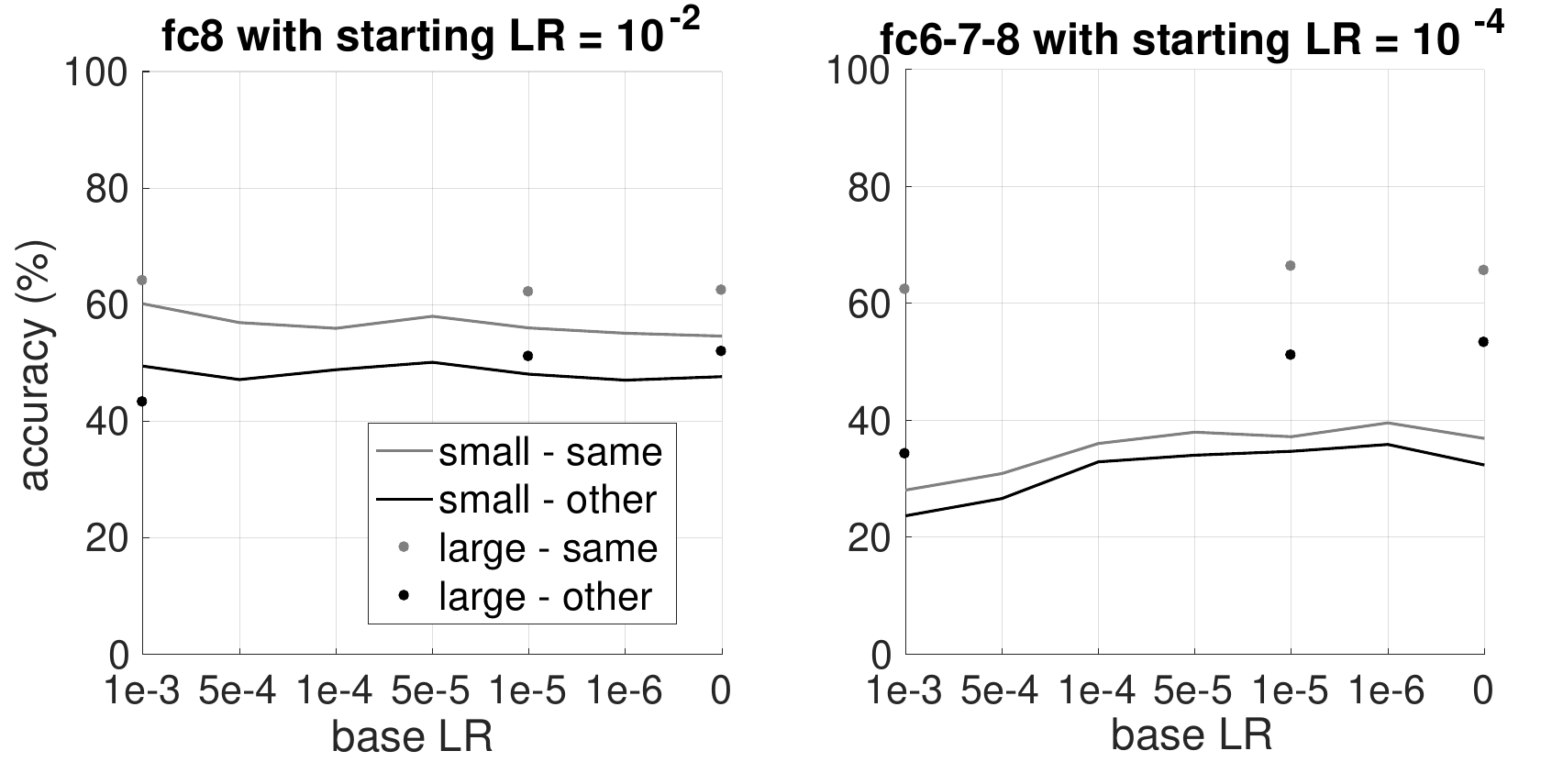}
\caption{Accuracy provided by  {\it CaffeNet} fine-tuned with different strategies: either learning from scratch only $fc8$ (Left) or $fc6, fc7$ and $fc8$ (Right). Performance is reported for both the same day of training (Light Gray) and a different one (Dark Gray). We tried multiple values of {\tt base LR} on the ``small'' training set (Continuous Line), and one small, medium and large values of {\tt base LR} (Dots) on the ``large'' one.}
\label{fig:presel-caffenet}
\end{figure}

We tried all possible combinations of values. For the parameters not mentioned, we kept their value as in {\sc Caffe} reference models.

We observed that the {\tt Dropout} percentage had a small influence and left it to the default value. We also observed that the polynomial {\tt LR Decay Policy} with $0.5$ slope consistently provided $5$-$10\%$ better accuracy. The most critical parameters proved being the {\tt Base LR} and the {\tt Learned FC Layers}. As an example, in Fig.~\ref{fig:presel-caffenet} we report the accuracy obtained respectively when learning only $fc8$ (Left) or $fc6, fc7$ and $fc8$ (Right). In both cases, we repeated fine-tuning with different {\tt Base LR} for all other layers, varied from $10^{-3}$ to $0$ (horizontal axis). Performance is reported, as in Fig.~\ref{fig:nystrom-same} and~\ref{fig:nystrom-full}, for both the same day of training (Light Gray) and a different one (Dark Gray). While we tried all values of {\tt Base LR} on the ``small'' experiment (Continuous Line), for the ``large'' experiment we limited to one small, medium and large value (Dots).

It can be observed that fine-tuning by learning from scratch only the last layer ($fc8$) is more robust to the small training set, for any {\tt Base LR} (Continuos Line in the range $40$-$60\%$), achieving best performance with higher {\tt Base LR} ($10^{-3}$). On the other hand, learning from scratch $fc6$-$7$-$8$ with small {\tt Base LR} ($10^{-5}$ or $0$) achieves best performance on the large training set. This is explained by noting that the three layers $fc6, fc7$ and $fc8$ involve a lot of parameters.

Based on these findings, we identified two representative strategies providing best performance respectively in small and large-scale settings: learning $fc8$ with large {\tt Base LR} ($10^{-3}$), that we call {\it adaptive} strategy, since it quickly adapts all layers to the new training set, and learning $fc6$-$7$-$8$ with {\tt Base LR} set to $0$, that we call the {\it conservative} strategy, since it slowly adapts only fully-connected layers. \\

 \begin{figure}[!t]
\centering
\includegraphics[width=\columnwidth]{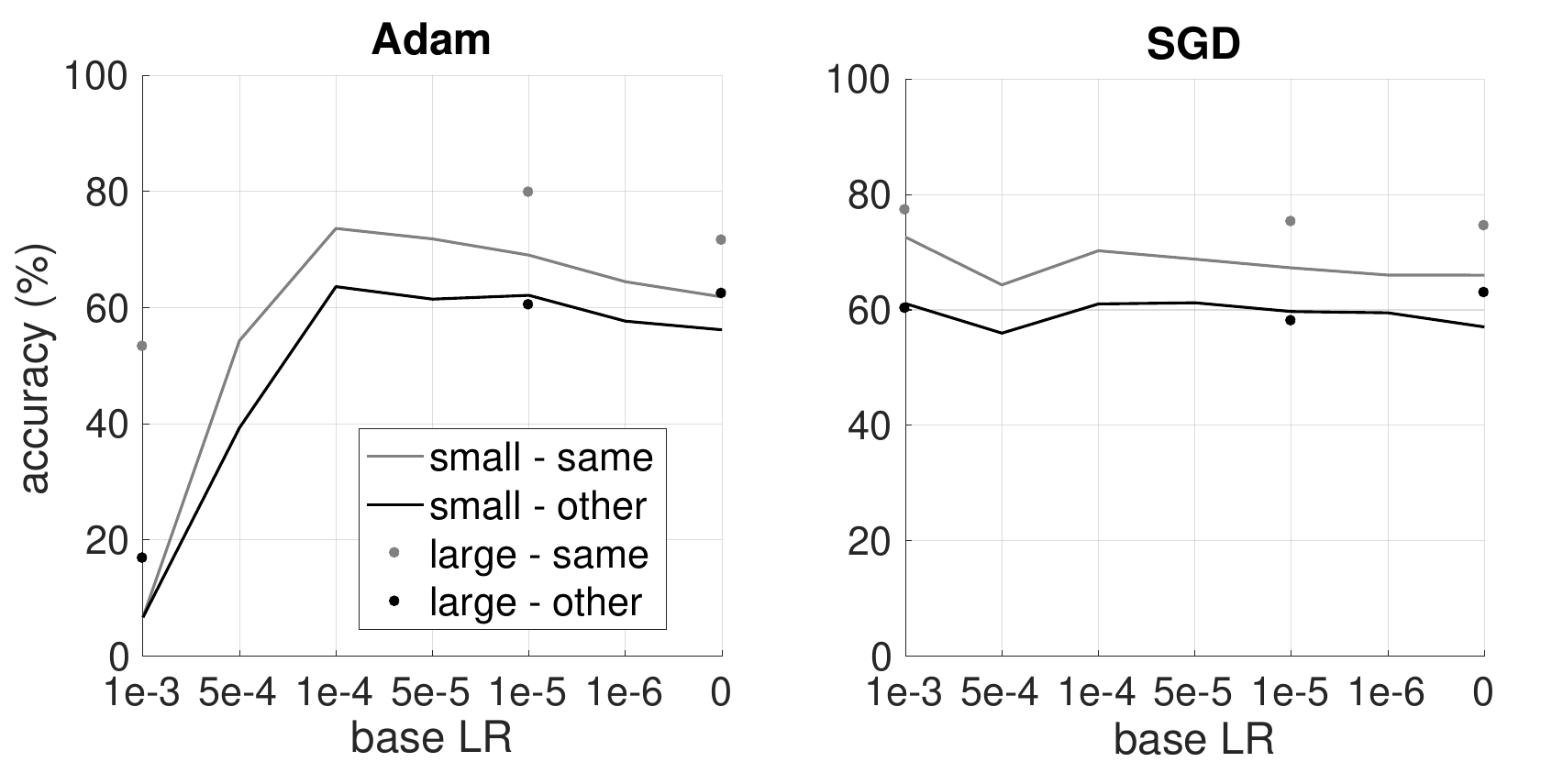}
\caption{Classification accuracy provided by fine-tuning {\it GoogLeNet} according to different strategies: either using the {\it Adam} solver (Left) or {\it SGD} (Right). The rest of the figure is similar to Fig.~\ref{fig:presel-caffenet}}
\label{fig:presel-googlenet}
\end{figure}

\noindent{ {\bf GoogLeNet.} } As explained in~\citep{szegedy2015}, this architecture is composed of a main branch, terminating with one FC layer ({\it loss3/classifier}), and two identical ``auxiliary'' branches terminating with two FC layers ({\it loss1(2)/fc}, {\it loss1(2)/classifier}). By considering this structure, we explored the following fine-tuning strategies:

\begin{itemize}
\item {\tt Base LR}: varied as for {\it CaffeNet}.
\item {\tt Learned FC Layers}: we always learned {\it loss3/classifier} from scratch with starting LR set to $10^{-2}$; regarding the auxiliary branches, we tried (i) to cut them out, (ii) to learn also {\it loss1(2)/classifier} from scratch with starting LR equal to $10^{-2}$, or, finally, (iii) to learn from scratch both {\it loss1(2)/fc} and {\it loss1(2)/classifier}, with starting LR set to $10^{-3}$.
\item {\tt Dropout}: we tried either the default {\sc Caffe} values ($40\%$ for {\it loss3/classifier} and $70\%$ for {\it loss1(2)/fc}) or $60\%$ for {\it loss3/classifier} and $80\%$ for {\it loss1(2)/fc}.
\item {\tt Solver}: we tried either {\it SGD}  or {\it Adam}~\cite{kingma2015} solvers in {\sc Caffe}.
\item {\tt LR Decay Policy}: when using {\it SGD}, we used polynomial decay with exponent $0.5$ or $-3$; when using {\it Adam}, we kept the learning rate constant.
\end{itemize}

As for {\it CaffeNet}, we tried all combinations and left other parameters to default {\sc Caffe} values. 

We first observed that for this architecture the impact of the {\tt Learned FC Layers} from scratch was small, with the three strategies behaving similarly. We chose the last one (iii), that was slightly more stable.
We also observed a little benefit from using higher {\tt Dropout} percentages. 

One critical aspect was instead the choice of the {\tt Solver}. To this end, in Fig.~\ref{fig:presel-googlenet} we report the accuracy achieved respectively when using {\it Adam} (Left) or {\it SGD} (Right). In the latter case, we applied the polynomial  {\tt LR Decay Policy} with $0.5$ exponent, since it was consistently better. Performance is reported, as in Fig.~\ref{fig:presel-caffenet}, for both the same day of training (Light Gray) and a different one (Dark Gray). We fine-tuned again with different {\tt Base LR} from $10^{-3}$ to $0$ (horizontal axis), trying all values on the ``small'' experiment (Continuous Line) and one small, medium and large value for the ``large'' experiment (Dots).

It can be observed that the {\it SGD} solver is more robust to variations of the {\tt Base LR}, but we opted for {\it Adam}, which provides better accuracies for mid-range values of {\tt Base LR}, both for the small and the large setting.
Similarly to {\it CaffeNet}, we identified an {\it adaptive} fine-tuning strategy with with {\tt Base LR} set to $0$ and a more {\it conservative} strategy with {\tt Base LR} set to $10^{-5}$.


 \begin{figure}[t]
 \centering
 \includegraphics[width=\columnwidth]{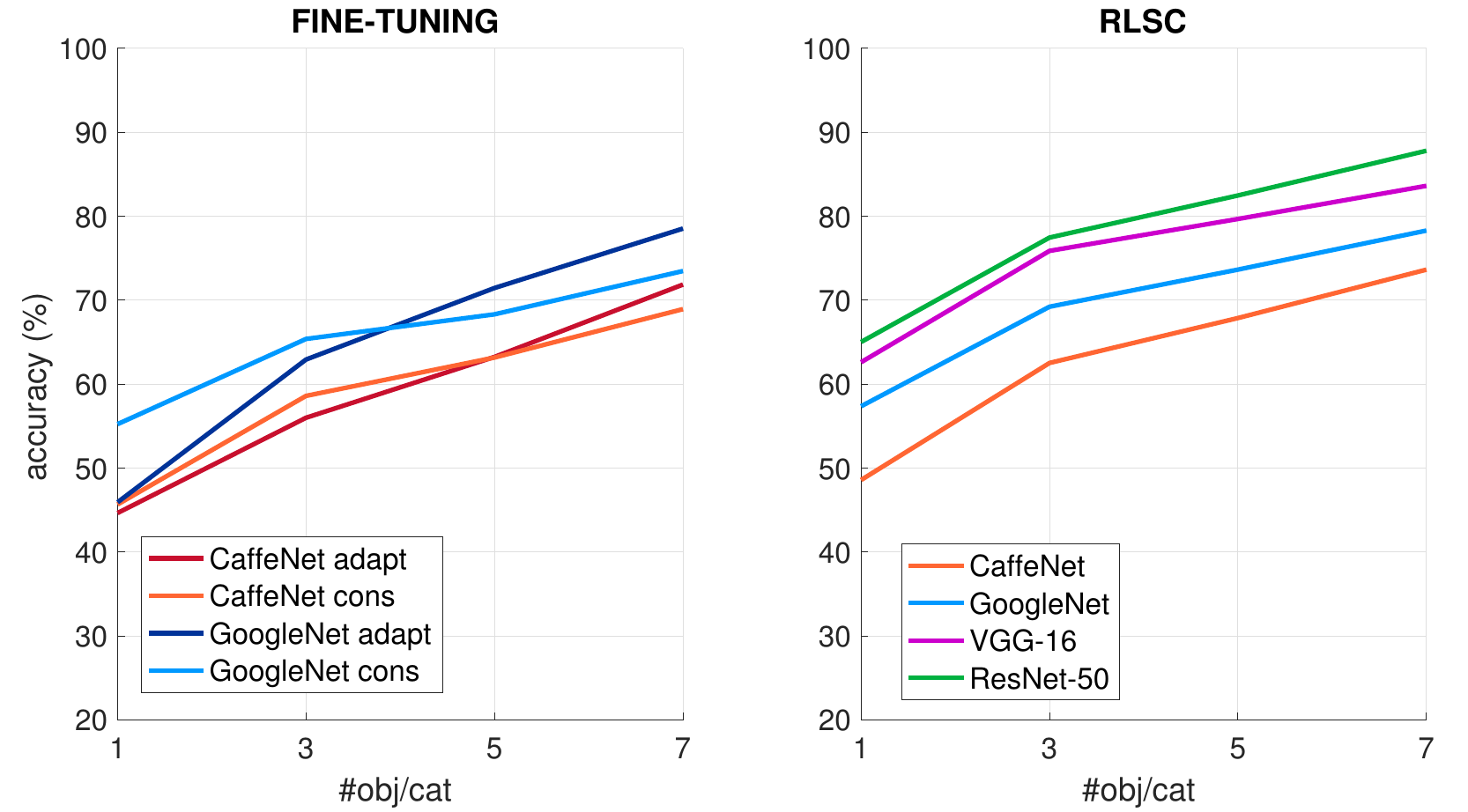}
 \caption{{\bf Recognition accuracy vs \# instances} (number of object instances available during training). Same experiment as Fig.~\ref{fig:semantic-variability} executed for a $10$-class categorization problem.}
 \label{fig:semantic-variability-10}
 \end{figure}

 \begin{figure}[t]
 \centering
 \includegraphics[width=\columnwidth]{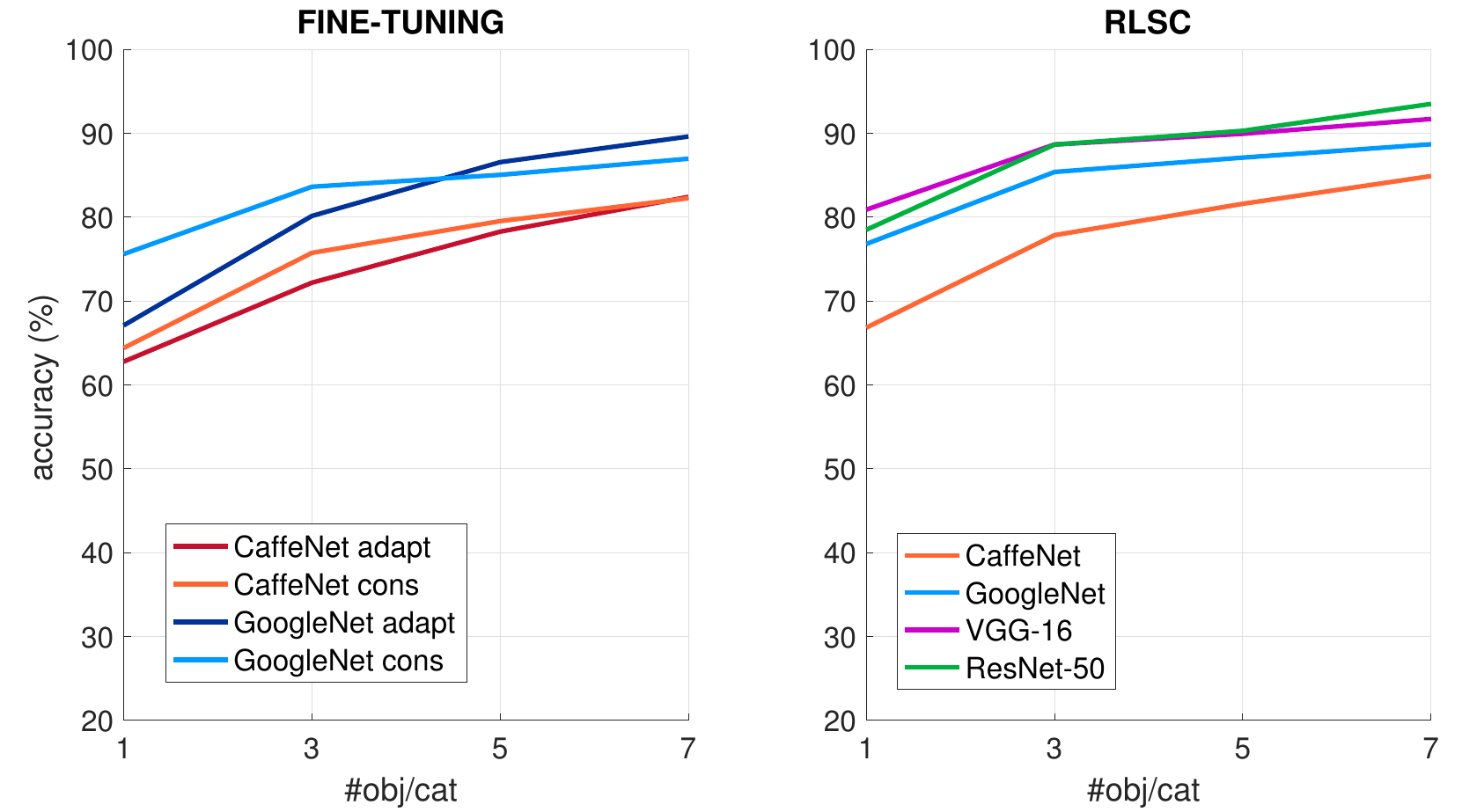}
 \caption{{\bf Recognition accuracy vs \# instances} (number of object instances available during training). Same experiment as Fig.~\ref{fig:semantic-variability} executed for a $5$-class categorization problem.}
 \label{fig:semantic-variability-5}
 \end{figure}
 
 \begin{figure}[t]
 \centering
 \includegraphics[width=\columnwidth]{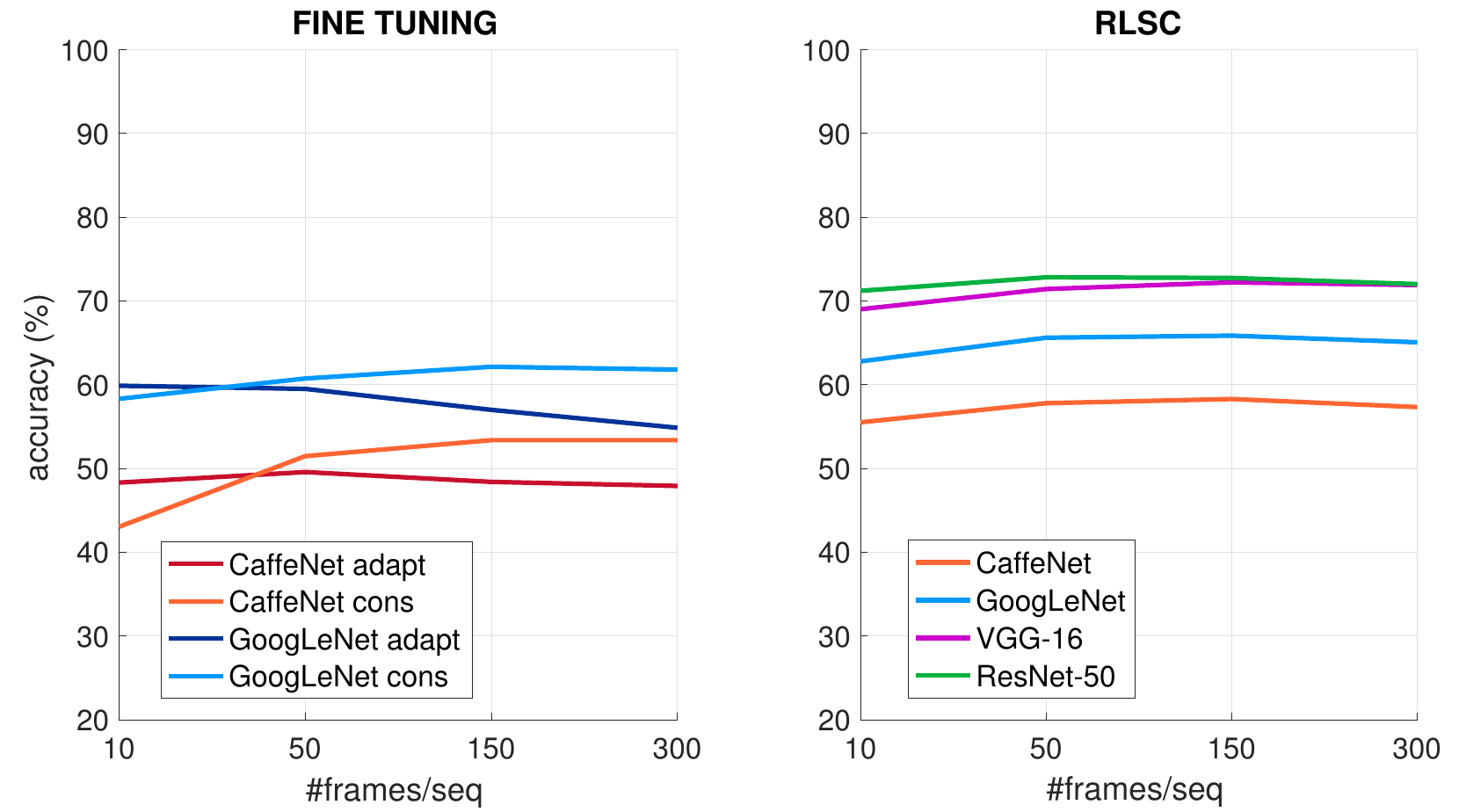}
 \caption{{\bf Recognition accuracy vs \# frames} (number of object views available during training). Same experiment as Fig.~\ref{fig:training-size-variability} but training only on $3$ object instances per category.}
 \label{fig:training-size-variability-3}
 \end{figure}
 
\section{Additional Experiments on Categorization}

In this Section we report additional results to show that the trends observed in the conditions considered in the paper hold also in other settings.

 \begin{figure*}[h]
 \centering
 \includegraphics[width=0.9\textwidth]{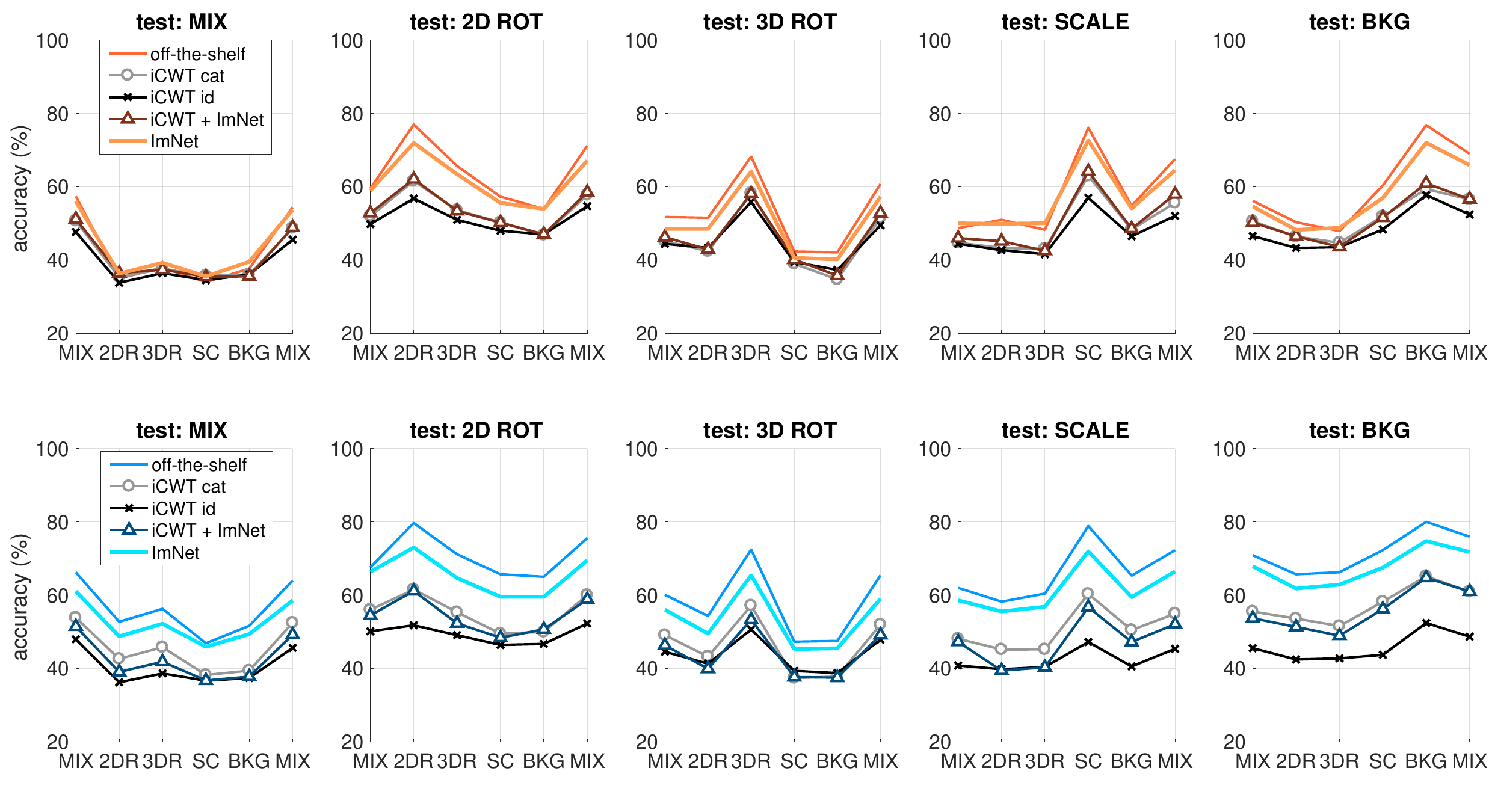}
 \caption{Same experimental setting as in Fig.~\ref{fig:invariance}: here we compare training RLSC on different image representations, provided by {\it CaffeNet} (Top, Orange) or {\it GoogLeNet} (Bottom, Blue) fine-tuned according to different strategies (see Sec.~\ref{sec:invariance-cat-adapted}).}
 \label{fig:invariance-cat-adapted}
 \end{figure*}
 
\subsection{What do we gain by adding more frames?}

Fig.~\ref{fig:training-size-variability-3} reports the experiment of Fig.~\ref{fig:training-size-variability}, when performed including only $3$ object instances per category in the training set. For completeness, here we provide the list of the considered $15$ categories: $cellphone$, $mouse$, {\it coffee mug}, {\it pencil case}, $perfume$, $remote$, {\it ring binder}, {\it soap dispenser}, $sunglasses$, $flower$, $wallet$, $glass$, $hairbrush$, {\it hair clip}, $book$.

This result confirms that adding more object views cannot be used as a viable strategy to improve categorization accuracy, which remains constant and definitely lower than the performance achieved with $7$ example instances per category (above $\sim 70\%$, see Fig.~\ref{fig:training-size-variability}).
 
\subsection{What do we gain by adding more instances?}
 
Fig.~\ref{fig:semantic-variability-10} and~\ref{fig:semantic-variability-5} report the experiment of Fig.~\ref{fig:semantic-variability}, when discriminating between respectively $10$ and $5$ object categories (rather than $15$). As it can be noticed, accuracy increases remarkably as more instances per category are made available, confirming that semantic variability is critical even in settings that involve few categories.
 
\subsection{Improving Viewpoint Invariance of CNNs}
\label{sec:invariance-cat-adapted}

As mentioned at the end of Sec.~\ref{sec:invariance-id-adapted}, here we investigate whether image representations from CNNs fine-tuned on subsets of iCWT can be better than off-the-shelf features also for categorization tasks. 

We repeat the $15$-class task of Fig.~\ref{fig:invariance}, performed by sampling $\sim 20$ frames per sequence, by training RLSC on features from the CNNs fine-tuned as in Sec.~\ref{sec:invariance-id-adapted}. Note that for ({\bf ImNet}) dataset we fine-tuned over the $15$ ImageNet synsets corresponding to the $15$ categories that involved in the categorization task. From the results reported in Fig.~\ref{fig:invariance-cat-adapted} (using the same notation as in Fig.~\ref{fig:invariance-id-adapt}), it can be clearly observed that none of these representations is better than off-the-shelf features for the categorization task.


\section{Generalization Across Days}
\label{sec:gen-across-days}

In this Section we test robustness to changes of illumination, background, etc., which are neither semantic nor geometric. The purpose of this material is to show how these aspects, albeit not in the scope of the paper, are very important and will be subject of future investigation.

We consider the common situation where the robot is asked to recognize an object that was showed him on a past day, possibly in a slightly different setting. We ask whether even small contextual variations (like a different time of day, or background configuration) can degrade the accuracy of the considered deep learning methods. 

To this end, while, in all experiments of the paper, training and testing is always performed on the same day of acquisition, in the following we report the performance drop experienced when testing the same models on a different day (we recall that in the current release of iCWT each sequence acquisition is repeated in two different days).\\

\begin{figure}[t]
\centering
\includegraphics[width=\columnwidth]{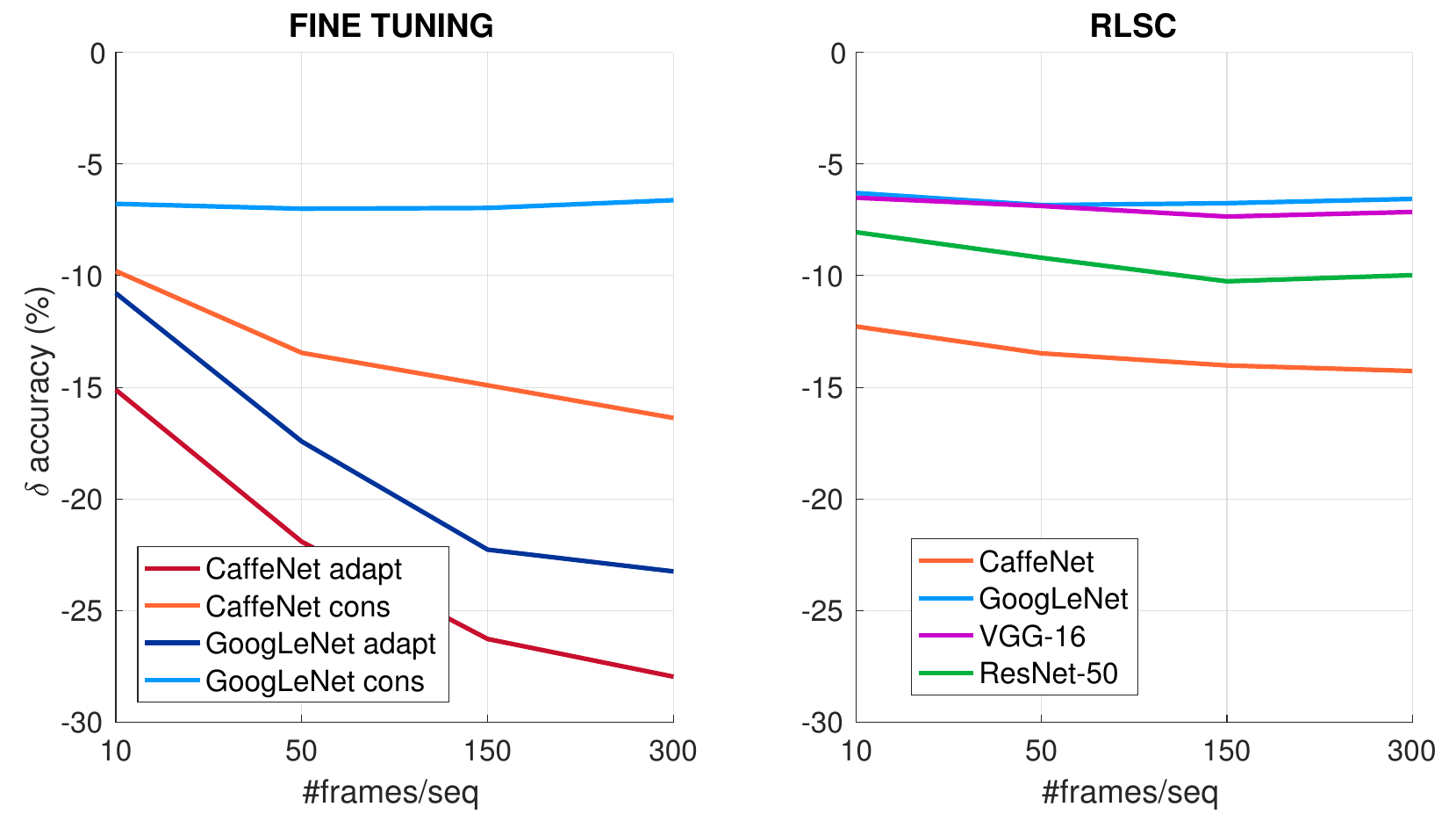}
\caption{{\bf Categorization accuracy vs \# frames - Generalization across days}. Drop in performance (difference between test accuracy) 
observed when testing the models trained for the categorization task reported in Sec.~\ref{sec:training-size-variability} on the same day of training and on a different one.}
\label{fig:training-size-variability-d12}
\end{figure}

\noindent{\bf Categorization.} Fig.~\ref{fig:training-size-variability-d12} report the performance drop observed respectively for the experiment of Fig.~\ref{fig:training-size-variability} (addition of example frames) and Fig.~\ref{fig:semantic-variability} (addition of object instances). The drop is computed as the difference between the accuracy reported in Fig.~\ref{fig:training-size-variability} (Fig.~\ref{fig:semantic-variability}) and the accuracy achieved when testing the models on a different day.

In both experiments, performance degrade (from $5\%$ up to almost $30\%$ for the {\it adaptive} fine-tuning). A larger drop for fine-tuning ({\it adaptive} in particular) suggests that this approach is more prone to overfitting the training day.
On the contrary, the $\sim5\%$ drop experienced when using less aggressive strategies as RLSC suggests that features from off-the-shelf networks as {\it GoogleNet} and {\it ResNet-$50$} can be quite robust.

\begin{figure}[t]
\centering
\includegraphics[width=\columnwidth]{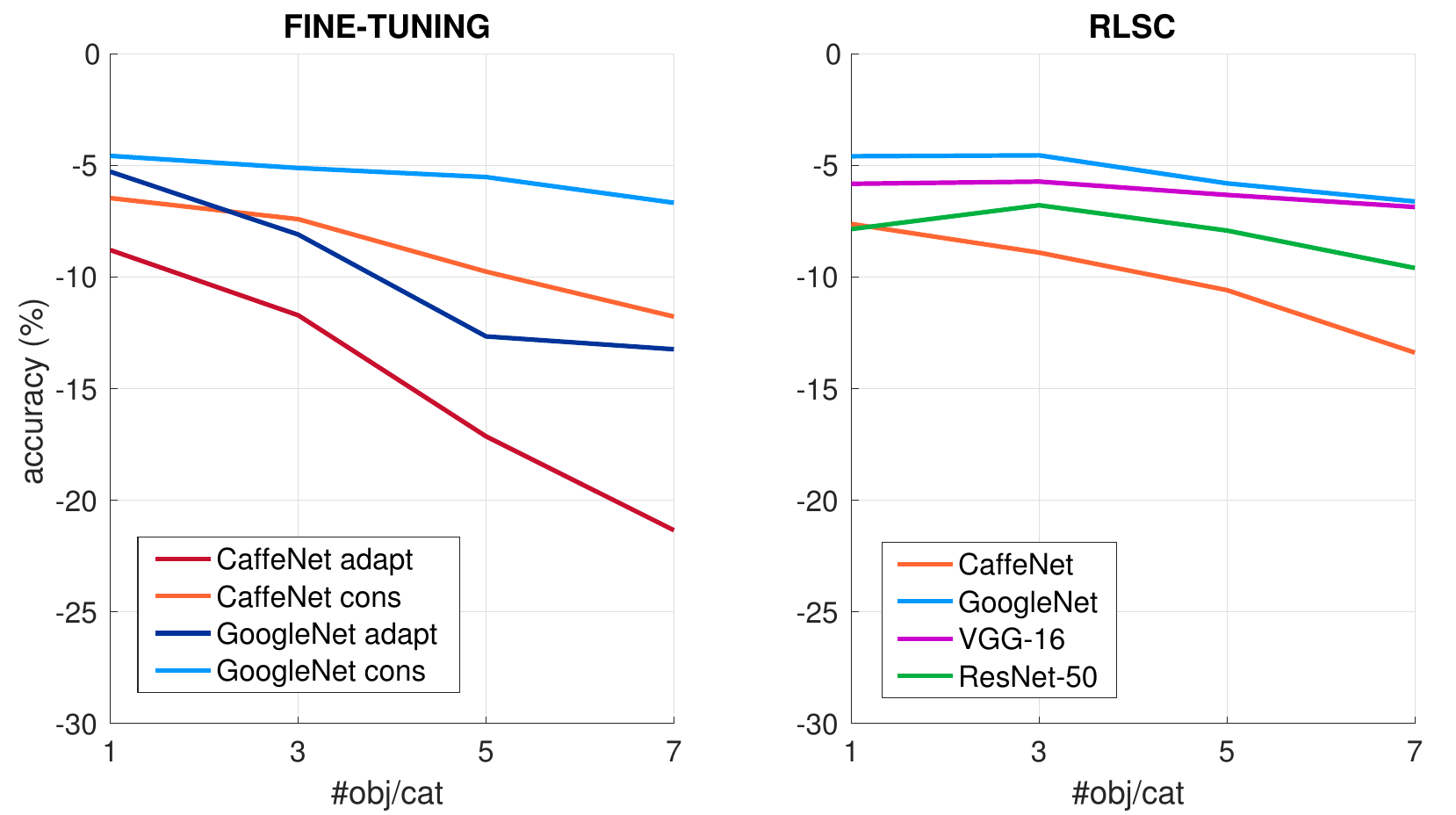}
\caption{{\bf Categorization accuracy vs \# instances - Generalization across days}. Drop in performance (difference between test accuracy) observed when testing the models trained for the categorization task reported in Sec.~\ref{sec:semantic-variability} on the same day of training and on a different one.}
\label{fig:semantic-variability-d12}
\end{figure}

Note that, when adding more example views, accuracy does not improve on the training day and even degrade on another day (Fig.~\ref{fig:training-size-variability-d12} and Fig.~\ref{fig:training-size-variability}). Differently, when adding more example instances, accuracy increases in both cases, but more on the training day than on another one (Fig.~\ref{fig:semantic-variability-d12} and Fig.~\ref{fig:semantic-variability}).\\

\begin{figure}[t]
\centering
\includegraphics[width=\columnwidth]{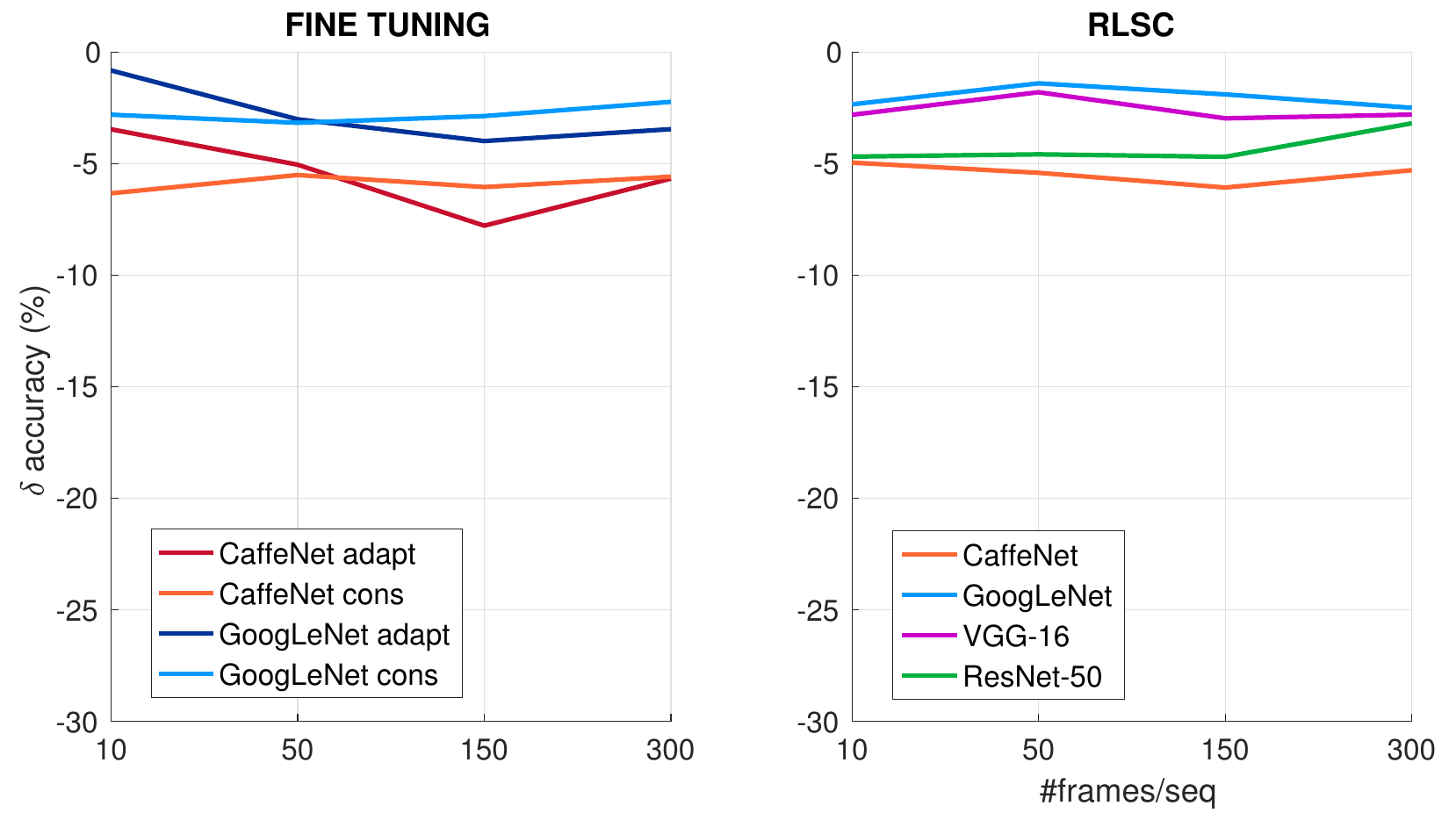}
\caption{ {\bf Identification accuracy vs \# frames - Generalization across days}. Drop in performance (difference between test accuracy) 
observed when testing the models trained for the identification task reported in Sec.~\ref{sec:training-size-variability-id} on the same day of training and on a different one.}
\label{fig:training-size-variability-d12-id}
\end{figure}

\noindent{\bf Identification.} Differently, in this setting {\em adaptive} fine-tuning does not exhibit a similar dramatic drop: in Fig.~\ref{fig:training-size-variability-d12-id} the drop with respect to Fig.~\ref{fig:training-size-variability-id} is small for all methods and does not increase by adding example frames. This positively indicates that adding example views in identification does not overfit the training day, and equally improves performance also on a different day.

\end{document}